\documentclass[runningheads]{llncs}

\usepackage{graphicx}
\usepackage{comment}
\usepackage{amsmath,amssymb}
\usepackage{color}
\usepackage{url}
\usepackage{hyperref}

\usepackage{graphicx}
\usepackage{booktabs}
\usepackage{enumitem}  
\usepackage[skip=7pt]{caption} 
\usepackage{subcaption}
\usepackage{mathtools}
\usepackage[dvipsnames, table]{xcolor}
\usepackage{arydshln}
\usepackage{tabularx}
\usepackage{multirow}
\usepackage{colortbl}
\definecolor{Gray}{gray}{0.9}
\definecolor{Gray2}{gray}{0.8}

\newif\ifreview
\reviewtrue
\reviewfalse

\ifreview
	\usepackage{lineno}

	\linenumbers
\fi

\begin{document}


\def\SubNumber{19}

\def\GCPRTrack{Young Researcher's Forum}

\title{Analytical Uncertainty-Based Loss Weighting in Multi-Task Learning}

\ifreview
\else

    \author{Lukas Kirchdorfer$^\ddagger$\inst{1,3} \and 
    Cathrin Elich$^\ddagger$ \inst{2,3}\and 
    Simon Kutsche$^\ddagger$ \inst{3}\and 
    Heiner Stuckenschmidt\inst{1}\and 
    Lukas Schott$^*$\inst{3} \and
    Jan M. Köhler$^*$\inst{3}
    }
    \authorrunning{L. Kirchdorfer et al.} 
    
    \institute{University of Mannheim, Germany
    \email{firstname.lastname@uni-mannheim.de} \and
    MPI for Intelligent Systems, Tübingen 
    \email{cathrin.elich@tuebingen.mpg.de} \and
    Center for AI, Robert Bosch, Germany
    \email{firstname.lastname@de.bosch.com} 
    \vspace{-0.25cm}
    {\centering \center \small  $^\ddagger$Work done during an internship/thesis at Bosch.\; \; \; $^*$Joint senior authors.
    }
    }

\fi

\maketitle              

\begin{abstract}
With the rise of neural networks in various domains, multi-task learning (MTL) gained significant relevance. A key challenge in MTL is balancing individual task losses during neural network training to improve performance and efficiency through knowledge sharing across tasks. 
To address these challenges, we propose a novel task-weighting method by building on the most prevalent approach of Uncertainty Weighting and computing analytically optimal uncertainty-based weights, normalized by a softmax function with tunable temperature. Our approach yields comparable results to the combinatorially prohibitive, brute-force approach of Scalarization while offering a more cost-effective yet high-performing alternative.
We conduct an extensive benchmark on various datasets and architectures. Our method consistently outperforms six other common weighting methods. Furthermore, we report noteworthy experimental findings for the practical application of MTL. For example, larger networks diminish the influence of weighting methods, and tuning the weight decay has a low impact compared to the learning rate.


  \keywords{Multi-Task Learning \and Deep Learning \and Computer Vision}
\end{abstract}
\section{Introduction} 
\label{sec:intro}
Multi-task learning (MTL) aims at solving multiple tasks simultaneously in a mutually beneficial manner. Intuitively, related tasks should share their knowledge and unrelated tasks should be processed more isolated to efficiently use the available data and compute resources. 

One of the major challenges is to find the right (implicit or explicit) balance between individual tasks to gain a good performance across tasks.
Recent approaches in deep learning tackle this problem from various angles, such as adapting the network architecture \cite{hu2021unit,heuer2021multitask} or resolving conflicts between task-specific gradients during optimization \cite{liu2021conflict,yu2020gradient}. 
A methodologically simple yet effective approach is based on providing task-specific weights during optimization, usually termed as \emph{loss weighting}.

Not explicitly weighting tasks, also termed \textit{Equal Weighting (EW)}, has potential pitfalls. Different tasks could be measured 
with different loss functions, such as L1 loss and cross entropy loss, which could lead to different loss scalings across tasks. Furthermore, some tasks might be more difficult than others and require more resources. Additionally, even when using the same loss function, variations in noisy data and prediction uncertainty contribute to differing loss magnitudes across tasks, necessitating the weighting of task losses. Thus, there have been various proposals for specialized weighting methods that dynamically adjust the weights along the course of training \cite{liu2021towards,chen2020just,javaloy2022rotograd,yu2020gradient,liu2021conflict,kendall2018multi,lin2022rlw,chennupati2019multinet++,liu2019end,chen2018gradnorm}. Despite reported improvements, there are inconsistencies within the MTL literature with regard to the ranking of different weighting methods, preventing practitioners
from making an informed decision about the choice of the weighting method for their problem at hand. This can lead to an inappropriate choice and harm the model's performance. Furthermore, Xin et al.\cite{xin2022current} experimentally show that most MTL methods are on a Pareto front w.r.t.\ their performance metrics and can be replaced by a merely brute force grid search, called \textit{Scalarization}, for optimal fixed weights. 

Our proposed method builds on the most prominent dynamic loss weighting method called Uncertainty Weighting (UW) \cite{kendall2018multi}, which learns homoscedastic uncertainty to weight tasks. 
UW, however, can be affected by bad initialization and is prone to overfitting.
To address those limitations, we adapt UW and derive the analytically minimal uncertainty weighting, which turns out to be the inverse of the losses. We then normalize these weightings with a softmax function with temperature parameter, which yields our method \textit{Soft Optimal Uncertainty Weighting} (UW-SO). 

We experimentally confirm the strong performance of Scalarization, however, it comes with high computational cost due to its combinatorial search of weights and is, therefore, inapplicable in many practical scenarios. Our approach UW-SO has only a single hyperparameter that can be optimized with significantly fewer steps while yielding comparable or better results than Scalarization.
Our experiments further reveal that larger architectures diminish the performance gain of weighting methods, especially Scalarization. 

We run an extensive set of experiments across various loss weighting methods with different datasets and architectures. The comparability and validity of the results are guaranteed due to method-specific tuning of learning rate (LR) and weight decay (WD). In accordance with \cite{xin2022current}, our results confirm that the absence of method-specific LR tuning is the major cause of ranking inconsistencies in previous works. Other causes are overfitting due to reporting validation instead of test data results as well as varying architectures and optimizers.

Overall, our contributions can be summarized as follows:
\begin{itemize}
\item We propose a novel loss weighting method termed UW-SO based on UW. 
\item We investigate the limitations of UW, such as overfitting and inertia which are the reasons why UW-SO consistently outperforms UW. 
\item We perform an extensive benchmark with tuned WD and LR to compare common loss weighting methods, demonstrating the superiority of UW-SO.
\item We provide noteworthy observations for the practitioner, for example, larger networks reduce the performance difference between loss weighting methods and LR tuning is more important than WD tuning.
\end{itemize}
\section{Related Work}
\label{sec:rel_work}
MTL considers training of tasks simultaneously by efficiently distributing resources and sharing knowledge between them~\cite{caruana1997multitask,ruder2017overview,vandenhende2021multi}.
There are typically two orthogonal research directions found in the recent literature:
One line of work considers multi-task architectures which focus on \textit{how features can be shared} between tasks~\cite{duong2015low,yang2016trace,misra_crossstitch_2016,liu2019end,xu_padnet_2018,maninis2019attentive}.
In this work, we make use of the basic hard-parameter sharing network structure, which consists of a fully-shared backbone and task-specific heads. We expect our method to be combinable with any other MTL network architecture.
In contrast, MTO methods aim to \textit{balance the tasks} to tackle the negative transfer that might occur between them. These methods can further be distinguished as gradient-based and loss weighting methods:

\textbf{Loss weighting methods} address the challenge of weighting task-specific losses appropriately. 
%
Most relevant for our work, Uncertainty Weighting (\textbf{UW}) \cite{kendall2018multi} weights different losses by learning the respective task-specific homoscedastic uncertainty. We adapt this by computing task weights based on the analytically optimal solution of UW and normalizing the results through a softmax function (Section~\ref{sec_uwso}). 
Lin et al.~\cite{lin2022rlw} argue that random sampling of loss weights (\textbf{RLW}) should be considered a relevant baseline.
Alternatively to the weighted sum of losses, the geometric loss strategy (\textbf{GLS})\cite{chennupati2019multinet++} computes the geometric mean.
While this method does not require any additional hyperparameters, it is numerically sensitive to a large number of tasks. 
Dynamic weight averaging (\textbf{DWA}) \cite{liu2019end} 
assigns a higher weight to tasks whose respective loss shows a slower decrease compared to other task losses.
Impartial Multi-Task Learning (\textbf{IMTL\hbox{-}L}) \cite{liu2021towards}
learns the scaling factor of the losses via gradient descent such that scaled losses would become constant for all tasks.
The brute force method \textbf{Scalarization} \cite{xin2022current} which searches all possible combinations of fixed loss weights 
has shown competitive performance compared to current automated MTO methods. Other loss weighting methods are proposed in~\cite{LakkapragadaSSW23,lin2023dual,Fan2022MaxGNR,guo_dtp_18}.
%
%

\textbf{Gradient-based methods} make direct use of task-specific gradients to either determine individual scaling factors which are applied on the task-wise gradients directly~\cite{chen2018gradnorm,sener2018multi,liu2021towards,navon2022multi,mao2022metaweighting,senushkin_alignedmtl_2023} or perform manipulations on the gradients to resolve potential alignment conflicts between them~\cite{liu2021towards,chen2020just,javaloy2022rotograd,shi_recon_2023,liu2023FAMO}. 
They mostly differ in the type of strategy used to handle these conflicts, such as projecting conflicting gradients onto the normal plane \cite{yu2020gradient}, or considering trade-offs between average and worst-case losses \cite{liu2021conflict}. 
A disadvantage of these methods is that computing task-wise gradients is computationally expensive.
In this paper, we do not consider gradient-based optimization methods as they have been shown to not outperform the simple loss weighting Scalarization approach \cite{xin2022current}, and Kurin et al. \cite{kurin2022defense} report that loss weighting methods commonly have significantly shorter training times \cite{chen2020just} which is relevant in practice.


\section{Background and Method}
\label{sec:method}
In MTL, we aim to resolve $K$ tasks for some input data point $x \in \mathcal{X}$.
For this, $x$ is mapped to labels $\{y_k \in \mathcal{Y}_k\}_{k\in [1,K]}$ simultaneously using specific mappings $\{f_k : \mathcal{X} \rightarrow \mathcal{Y}_k\}$. 
We assume hard task-shared parameters $\theta$ in a hydra-like neural network architecture.
This means all tasks receive the same intermediate feature $z = f (x; \theta)$ from the shared backbone and each task head yields output $f_k (x) = f'_k (z; \theta_k)$ with task-specific parameters $\theta_k$  \cite{ruder2017overview}. 

The network is trained by considering all tasks' losses $L_k$.
Naively summing up these losses (the equal weighting method) typically leads to imbalanced learning as tasks with high loss magnitude might dominate the training.
The goal is thus to find optimal (dynamic) loss weights $\omega_k$ for all tasks to optimize the loss $L=\sum_k \omega_k L_k$ in a way that tasks benefit w.r.t. their final performance metrics. 

\subsection{Weaknesses of Uncertainty Weighting and Scalarization}
\label{sec:weak_uw}
\begin{figure}[t]
     \centering
     \begin{subfigure}[t]{0.3\textwidth}
         \centering
         \includegraphics[trim = 0mm 0mm 0mm 0mm, clip, width=\textwidth]{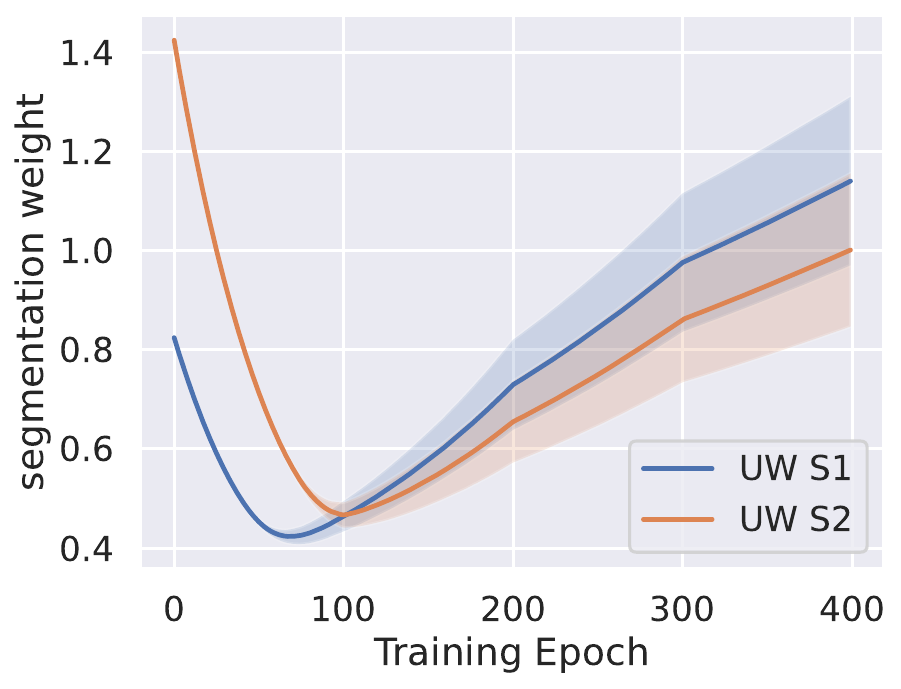}
         \caption{Segmentation}
     \end{subfigure}
     \hfill
     \begin{subfigure}[t]{0.3\textwidth}
         \centering
         \includegraphics[trim = 0mm 0mm 0mm 0mm, clip, width=\textwidth]{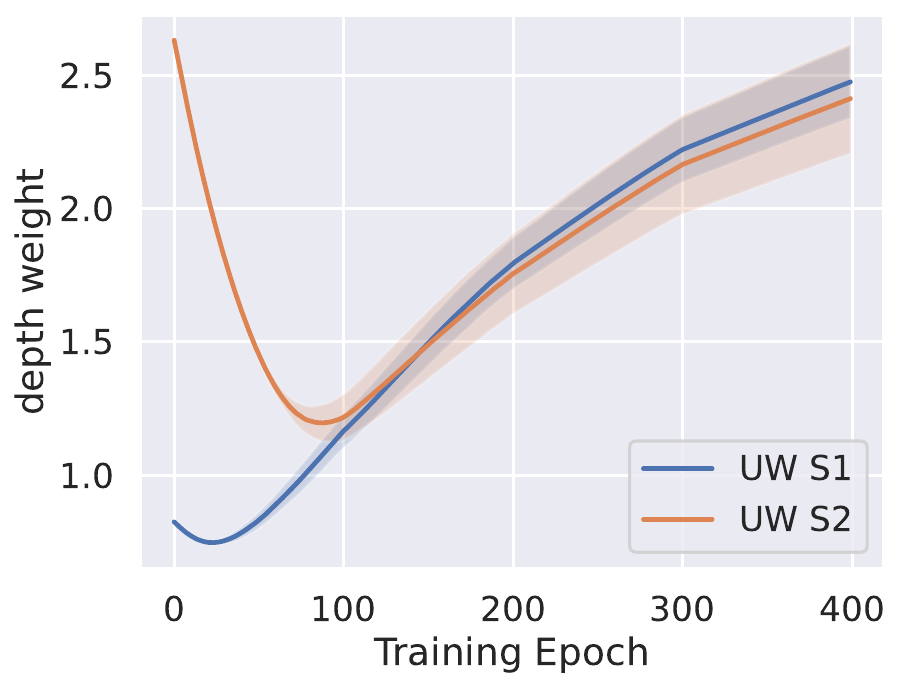}
         \caption{Depth}
     \end{subfigure}
     \hfill
    \begin{subfigure}[t]{0.3\textwidth}
         \centering
         \includegraphics[trim = 0mm 0mm 0mm 0mm, clip, width=\textwidth]{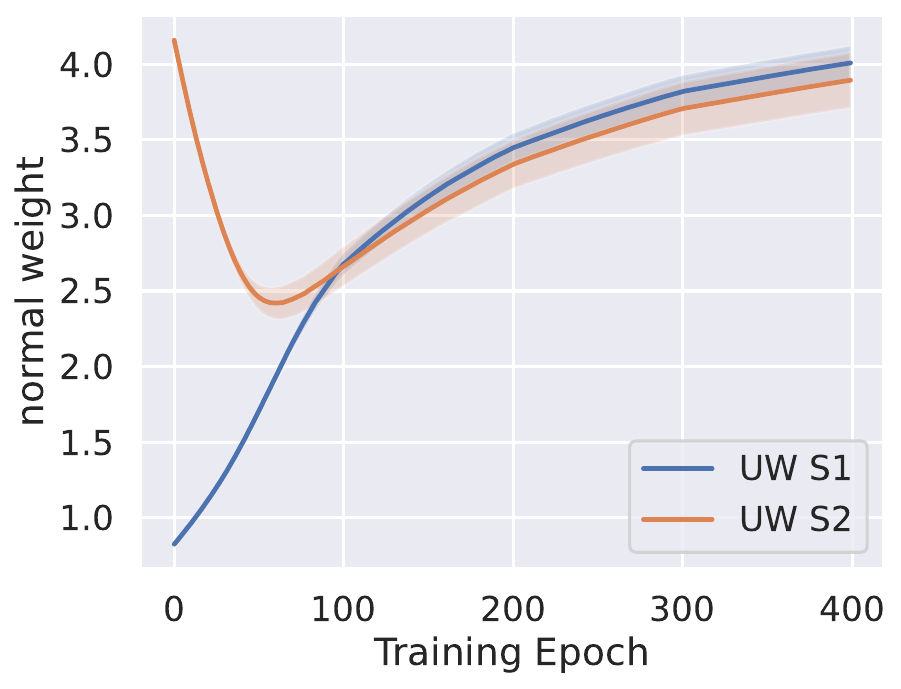}
         \caption{Normal}
     \end{subfigure}
     \hfill
    \caption{Comparison of the learning procedure of task weights for a) semantic segmentation, b) depth estimation, and c) surface normals on NYUv2 using SegNet for two different initializations of $\sigma_t$ for UW. Equal starting parameter values in blue (UW S1) as in \cite{LibMTL}; higher starting values (values of last epoch from a previous run) in orange (UW S2). The plots do not show $\sigma_t$ values, but actual task weights $\omega_t = \frac{1}{2\sigma_t^2}$. 
    We plot the mean task weight of 5 random seeds with the standard deviation as shaded area.
    }
    \label{fig:weakness_UW}
\end{figure}

\textbf{UW} \cite{kendall2018multi} is one famous MTO approach with over 3.2k citations (May'24) 
and yields competitive performance (see Section \ref{sec:comparison_common_lw_methods}) besides its simplicity.
However, UW also shows some drawbacks: 
First, we observe that UW can be affected by \textit{bad initialization / inertia}. 
As uncertainty weights are usually initialized equally for all tasks, it can slow down their progress toward reaching the best values for each task and epoch using gradient descent, especially as task weights often differ in orders of magnitude. We refer to this phenomenon as \textit{update inertia}. 
To demonstrate this phenomenon empirically, we focus on the development of task weights $\omega_t$ of the NYUv2 dataset with two different initializations of $\omega_t$  (Figure \ref{fig:weakness_UW}). In the first initialization setting (blue line, UW S1), we consider the usual initialization of $\omega_k=0.8$~\cite{LibMTL}. In the other setting (orange line, UW S2), we initialize the task weights higher and choose the initialization values equal to the final $\omega_t$ of a previous run.  
We observe that the learned task weights develop differently due to the different initializations. It takes roughly 100 epochs (1/4 of the whole training) to recover, i.e., the blue and the orange line then behave similar. As both experiments receive the same non-weighted losses from each task at the beginning we would expect that its weightings $\omega_k$ adapt quickly to the task loss ignoring the wrong initialization value. 
Thus, a non-optimal initialization has a direct impact on the training dynamics due to the update inertia of the task weights. 
Another example for \textit{update inertia} is discussed in section~\ref{sec:appx_weak_uw}  for the CelebA dataset.

Second, we observe that UW is \textit{prone to overfitting} - a detailed discussion is presented after the benchmark results in Section~\ref{sec:comparison_common_lw_methods}.

Lastly, we give a first hint with a toy example in Section \ref{sec:appx_weak_uw} that UW does not only depend on task-wise aleatoric homoscedastic uncertainty but also shows a \textit{model complexity dependence}.

\textbf{Scalarization} is demonstrated \cite{xin2022current} to yield competitive performance on a range of MTL problems, but there are shortcomings:
As mentioned by the authors, manually tuning loss weights by performing an extensive grid search is \textit{computationally expensive}.
Besides, it can only be applied to problems with a small number of tasks, as the \textit{combinatorial complexity} becomes prohibitive. With three tasks and a step size of 0.1, there are a total of 36 distinct weight combinations to consider (Figure \ref{fig:parallel_coord_scalar} visualizes these combinations for three different LRs, thus 108 experiments and only three of those improve upon Single-Task Learning (\textbf{STL})).
Furthermore, the \textit{choice of the granularity} of the search space is not obvious. In our experiments, we find high discrepancies in sensitivity regarding the needed granularity for the datasets. 
Besides, it is also argued that \textit{adaptive loss weights} can be beneficial~\cite{guo_dtp_18}.

\subsection{Our contribution: Soft Optimal Uncertainty Weighting}
\label{sec_uwso}

AS UW \cite{kendall2018multi} shows inertia with task uncertainty $\sigma_k$ being updated gradually through gradient descent, we are determining which $\sigma_k$ values would analytically minimize the total Loss $L$ in a given batch. These values are then normalized using a softmax function with temperature.

\paragraph{UW-O: Minimizing the total loss in UW}
The approach UW \cite{kendall2018multi} weights losses based on their task-specific homoscedastic aleatoric uncertainty. The exact weighting formulae depend on the type of loss. For instance, for tasks with an $L_1$ loss it can be derived by assuming a Laplace posterior distribution and identifying $\sigma_k$ with the uncertainty of each task $k \in \mathcal{K}$, treating them as learnable parameters that are input independent
\begin{equation}
    L = \sum_{k \in \mathcal{K}}  \frac{1}{\sigma_k} L_k + \log \sigma_k. 
    \label{ew:uw}
\end{equation}
Here, $L_k$ is the task-specific loss, e.g., a mean absolute error. Intuitively, the $\sigma_k$ in the first term allows to down-weight difficult tasks by increasing the uncertainty represented by $\sigma_k$. The second term acts as a regularizer and prohibits the trivial solution of $\sigma_k \to \infty$.

Instead of learning $\boldsymbol{\sigma}$, we derive the analytically optimal solution and show its derivation for $L_1$ loss (see Section \ref{sec:appx_derivation_uwso} for a derivation for the $L_2$ and Cross-Entropy loss).

The objective in UW for tasks with an $L_1$ loss can be formulated as $\min_{\sigma_k} L$  \cite{kendall2018multi},
where we minimize the UW loss function with respect to $\sigma_k$.
Taking the derivative and solving for $\sigma_k$ results in an analytically optimal solution:
%
%
\begin{equation}
    \sigma_k = L_k
    \label{formula:UW-O_sigma}
\end{equation}

Replacing $\sigma_k$ with its analytical solution $L_k$ in the total loss function (Eq.~\ref{ew:uw}) gives the following loss: $L = \sum_{k \in \mathcal{K}}\frac{1}{sg[L_k]} L_k + \log sg[L_k],$
where we denote $sg$ as the stopgradient operator to avoid zero gradients of the network updates. Since we do not compute any gradient of the second part of the loss, we can simplify the term, such that

\begin{equation}
    L = \sum_{k \in \mathcal{K}} \frac{1}{sg[L_k]} L_k.
    \label{formula:UW-O}
\end{equation}

For later reference, we name this intermediate result Optimal Uncertainty Weighting (UW-O), where optimal refers to the analytical loss minimum. Interestingly, we found that there exist three approaches that lead to a similar solution: 1) \textbf{IMTL-L} \cite{liu2021towards} aims to have each weighted loss $\omega_k L_k$ scaled to $1$, though, they learn $\omega_k$ using gradient descent. 
2) \textbf{Dual-balancing} \cite{lin2023dual} transforms the loss to $\log L$ to normalize over different scales. Taking the gradient of $\log L$ is equivalent to taking the gradient of $\frac{1}{sg[L_k]} L_k$  \cite[Sec. 3.1] {lin2023dual} which is 1 for all tasks $k$; for the $L=||\cdot||_1$ loss it is equivalent to equation~\ref{formula:UW-O}. Thus, in dual-balancing the gradient is scaled whereas we scale the loss $L$.
3) \textbf{EMA} \cite{LakkapragadaSSW23} scale the loss by the Exponential Moving Average which is identical to the Inverse Loss when the hyperparameter is $\beta=1$.

\paragraph{UW-SO: Soft Optimal Uncertainty Weighting}
To challenge the high-performing Scalarization approach, we propose UW-SO. Our experimental results confirm the effectiveness of simply tuning task weights that sum to 1, as done in Scalarization. However, to overcome the inherent computational cost dependency on the number of tasks, while simultaneously achieving strong results, we advocate the application of the tempered softmax function to the inverse loss scaling of UW-O (Eq.~\ref{formula:UW-O}), yielding the UW-SO weighting:

\begin{equation}
    L = \sum_{k \in \mathcal{K}} \frac{exp({{\frac{1}{sg[L_k]}}/T)}}{\sum_{j \in \mathcal{K}} exp({{\frac{1}{sg[L_j]}}/T)}} L_k,
\label{formula:UW-SO}
\end{equation}

where we denote $T$ as the softmax temperature, 
with a higher value of $T$ resulting in more evenly distributed task weights with $\sum_{k \in \mathcal{K}} \omega_k = 1$. Contrary to Scalarization which requires the tuning of $\mathcal{K}$ task weights, thus being infeasible for a large number of tasks, our method UW-SO requires only tuning of one hyperparameter $T$.

We have chosen the (tempered) softmax function as it is widely known and used. The MTL methods RLW and DWA also use it for normalization with $T=1$ and $T=2$, respectively, stating it to be common in this field. Other domains related to discrete selection also make use of the tempered softmax: \cite{caccia2019language} uses it to control the quality-diversity trade-off of generated samples in GANs. \cite{hinton2015distilling} use it in knowledge distillation to provide softer outputs from the teacher model. Other usages are out-of-distribution detection \cite{liang2018enhancing} or confidence calibration \cite{guo2017calibration}. Works in hash learning \cite{tan2020learning} and for Neural-Architecture search \cite{chang2019differentiable} uses both the Gumbel-Softmax with temperature \cite{jang2016categorical}. 
\section{Experiments and Results}
\label{sec:exp}
\subsection{Experimental setup}

In this Section, we provide details about the used datasets, network architectures, metrics, and training proceeding.
Further details can be found in Section~\ref{sec:appx_impl_details}.

\label{sec:experimental_setup}
\textbf{Datasets}
We use three common computer vision MTL datasets: two datasets for scene understanding — \textit{NYUv2} \cite{Silberman:ECCV12} and \textit{Cityscapes}  \cite{Cordts2016Cityscapes} — and a binary attribute dataset \textit{CelebA} \cite{liu2015faceattributes}.
NYUv2 and Cityscapes comprise the tasks of semantic segmentation and depth estimation. The third task for NYUv2 is surface normals estimation.
CelebA constitutes a 40-class binary classification problem.

\textbf{Architectures} For NYUv2, we use a SegNet \cite{badrinarayanan2017segnet}, ImageNet pretrained ResNet-50 / ResNet-101 with a DeepLabHead, and the MTAN on top of the SegNet \cite{liu2019end}. For Cityscapes, we use a SegNet, a DeepLabV3+ \cite{chen2018encoder} network with pre-trained ResNet-50 / ResNet-101 backbones, and again the MTAN/SegNet. All Single-task learning (\textbf{STL}) baselines are trained with the SegNet. For CelebA, we use a ResNet-18, also for STL. 

\textbf{Metrics}
To compare models, we use task-specific metrics and the established $\Delta_m$-metric \cite{maninis2019attentive}. 
It measures the average relative performance gain of the multi-task model $M_m$ w.r.t.\ a single-task baseline $M_b$: $\Delta_m = \frac{1}{K}\sum_{k=1}^K(-1)^{l_k}(M_{m,k}-M_{b,k})/M_{b,k},$
where $l_k$ is $1 / 0 $ if a higher / lower value is better for criterion $M_k$. 

\textbf{Two Evaluation Setups}
Several papers (e.g., \cite{liu2021conflict,liu2019end,yu2020gradient,navon2022multi}) have used a fixed training protocol for NYUv2 and Cityscapes, with no hyperparameter tuning, averaging results over the last 10 test epochs. In contrast, other studies (e.g., \cite{xin2022current,kurin2022defense,sener2018multi}) advocate for method-specific hyperparameter tuning, which is more relevant for practitioners. Following \cite{xin2022current}, we perform a thorough hyperparameter search for all methods, selecting the best combination based on the $\Delta_m$ score and using early stopping on the validation set. For final evaluation, we train on 5 random seeds and report the mean test performance. However, acknowledging other works, we also provide results using the fixed protocol with MTAN/SegNet on NYUv2 and Cityscapes.

All models are trained with the Adam optimizer which has been shown to perform advantageous on MTL setups \cite{eli2023_challenging}.
Compared to \cite{liu2021conflict}, we increase the number of epochs for NYUv2 / Cityscapes to 400 / 600 epochs. CelebA remains at 100 epochs as there is no further improvement.

\textbf{Settings for MTO approaches}
For tuning the LR/WD and the hyperparameters of the MTO method, we employ a sequential line search by first tuning the LR together with the MTO hyperparameters using a fixed WD, followed by tuning the WD. 
For UW-SO, we search $T$ with a step size of 5 and then employ a finer search around the optimum. 
For Scalarization, we first test each possible combination of task weights with a step size of 0.1. If no proper result could be achieved (e.g., for Cityscapes), we further test values around the previously found optimum with a step size of 0.02. For DWA, we follow \cite{liu2019end} and set $T=2$. 
\newcolumntype{g}{>{\columncolor{Gray}}r}
\begin{table}[t]
\centering
\tiny
\setlength\tabcolsep{2pt} 
\newcommand{\gcs}{\hspace{4pt}}
\caption{Test data results on Cityscapes with SegNet, ResNet-50, ResNet-101, and MTAN. Our method UW-SO is underlined. For the first 3 architectures, we report the average over 5 runs, for $\Delta_m$ including $\pm$ one std. dev. The best score is in bold, the second best is underlined. We report the best LR and WD values for each experiment. WD and LR are abbreviated as: a: $0.0$, b: $10^{-6}$, c: $5\times10^{-6}$, d: $10^{-5}$, e: $5\times10^{-5}$, f: $10^{-4}$, g: $5\times10^{-4}$, h: $10^{-3}$, i: $5\times10^{-3}$, j: $10^{-2}$. The best softmax temperature for the 4 architectures is $T=20/28/48/22$. For Scalarization, best weights are equal for the first 3 architectures with $\omega = [0.02, 0.98]$ and for MTAN with $\omega = [0.04, 0.96]$ for segm. and depth. Additionally, we report results over 3 seeds based on the training protocol that uses MTAN+SegNet with fixed hyperparameters.
}
\begin{tabular}{cclll@{\gcs}ccc@{\gcs}ccc@{\gcs}cc@{\gcs}cg}
\toprule
               \multirow{3}{*}{A.} & \multirow{3}{*}{E.} & \multirow{3}{*}{Method} & \multirow{3}{*}{lr} & \multirow{3}{*}{wd} && \multicolumn{2}{c}{Segmentation $\uparrow$} && \multicolumn{2}{c}{Depth $\downarrow$} && \multirow{3}{*}{Ep} \\ 
                     & &  & & &&        mIoU & PixAcc &&        AbsErr & RelErr &&   && $\Delta_m\%\downarrow$\\
\midrule
 \multirow{10}{*}{\rotatebox[origin=c]{90}{SegNet}} & \multirow{10}{*}{\rotatebox[origin=c]{90}{early stopping}} & STL &g&f,d  &&         $0.723$ & $0.927$ &&        $0.0124$ & $24.5$  \\ \hline
  && Scalar & f & d &&  0.707 &  0.919 &&    \textbf{0.0123} &      \underline{25.7} && 531.2  &&      $\pmb{1.9 \pm 0.9}$ \\
  && EW & g & d &&  \textbf{0.731} &  \textbf{0.928} &&    0.0157 &      76.7 && 560.8  &&        $59.9 \pm 5.0$ \\
  && RLW&f&f &&  0.722 &  0.926 &&    0.0170 &     107.1 && 529.6  &&       $93.8 \pm 14.8$ \\
  && DWA&f&f &&  \underline{0.729} &  \underline{0.927} &&    0.0158 &      85.9 && 584.4  &&        $69.6 \pm 9.7$ \\
  && GLS&g&d &&  0.717 &  0.923 &&    0.0129 &      27.5 && 564.2  &&         $4.5 \pm 0.6$ \\
  && IMTL-L&h&d &&  0.715 &  0.922 &&    \underline{0.0128} &      33.5 && 568.6  &&         $10.6 \pm 3.6$ \\
  && UW&h&d &&  0.718 &  0.923 &&    \underline{0.0128} &      32.1 && 550.6  &&         $9.0 \pm 3.0$ \\
   && \underline{UW-SO} &g&d &&  0.674 &  0.907 &&    0.0130 &      \textbf{25.2} && 544.0  &&         $\underline{4.4 \pm 0.9}$ \\
\midrule
 \multirow{8}{*}{\rotatebox[origin=c]{90}{ResNet-50}}
  & \multirow{8}{*}{\rotatebox[origin=c]{90}{early stopping}} & Scalar&h&a &&  0.738 &  0.929 &&    0.0117 &      \underline{28.3} && 417.2  &&       $2.1 \pm 1.4$ \\
  &&EW&h&a &&  \underline{0.757} &  \textbf{0.936} &&    0.0119 &      31.9 && 448.4  &&       $5.3 \pm 0.7$ \\
  &&RLW&g&a &&  \underline{0.757} &  \textbf{0.936} &&    0.0119 &      32.7 && 486.6  &&        $6.1 \pm 1.1$ \\
  &&DWA&g&a &&  \textbf{0.758} &  \textbf{0.936} &&    0.0118 &      31.4 && 404.6  &&       $4.4 \pm 0.7$ \\
  &&GLS&h&b &&  0.755 &  0.934 &&    0.0115 &      29.1 && 495.8  &&        $1.7 \pm 1.2$ \\
  &&IMTL-L&g&a &&  0.754 &  \underline{0.935} &&    0.0113 &      28.6 && 350.8  &&       $0.7 \pm 0.9$ \\
  &&UW&h&a &&  0.752 &  0.934 &&    \underline{0.0113} &      \underline{28.3} && 415.0  &&        $\underline{0.5 \pm 0.8}$ \\
  &&\underline{UW-SO} &g&a &&  0.748 &  0.933 &&    \textbf{0.0112} &      \textbf{28.0} && 363.4  &&        $\pmb{0.3 \pm 1.3}$ \\
\midrule
 \multirow{8}{*}{\rotatebox[origin=c]{90}{ResNet-101}}
  & \multirow{8}{*}{\rotatebox[origin=c]{90}{early stopping}} & Scalar&f&d &&  0.740 &  0.931 &&    \underline{0.0115} &      31.6 && 462.4  &&       $4.8 \pm 4.7$ \\
  &&EW&h&a &&  0.749 &  0.933 &&    0.0121 &      32.8 && 427.0  &&       $7.0 \pm 0.7$ \\
  &&RLW&g&a &&  \underline{0.752} &  \underline{0.934} &&    0.0119 &      32.7 && 482.0  &&        $6.3 \pm 0.8$ \\
  &&DWA&g&b &&  \textbf{0.753} &  \textbf{0.935} &&    0.0118 &      32.2 && 457.6  &&        $5.6 \pm 0.9$ \\
  &&GLS&g&a &&  0.743 &  0.931 &&    0.0116 &      28.9 && 408.6  &&        $2.1 \pm 0.6$ \\
  &&IMTL-L&h&b && 0.743 &  0.931 &&    0.0116 &      \textbf{28.2} && 374.2  &&        $\underline{1.6 \pm 1.4}$ \\
  &&UW&h&b && 0.744 &  0.931 &&    0.0116 &      \underline{28.5} && 337.8  &&        $1.8 \pm 1.0$ \\
  &&\underline{UW-SO} &g&a &&  0.749 &  0.933 &&    \textbf{0.0113} &      28.7 && 357.6  &&        $\pmb{1.1 \pm 0.9}$ \\

\midrule
 \multirow{8}{*}{\rotatebox[origin=c]{90}{MTAN/SegNet}}

& \multirow{8}{*}{\rotatebox[origin=c]{90}{avg. last 10 ep.}} & Scalar &f&a &&  0.721 &  0.927 &&    \underline{0.0130} &      \underline{27.7} && avg &&   \underline{$-0.7 \pm (1.2)$} \\
&& EW     &f&a &&  0.743 &  0.932 &&    0.0158 &      44.4 && avg &&  $17.8 \pm (3.1)$ \\
&& RLW    &f&a &&  0.736 &  0.931 &&    0.0159 &      45.3 && avg &&  $19.1 \pm (1.8)$ \\
&& DWA    &f&a &&  0.743 &  \underline{0.933} &&    0.0159 &      45.6 && avg &&  $19.1 \pm (2.7)$ \\
&& GLS    &f&a &&  0.729 &  0.930 &&    0.0136 &      29.7 && avg &&   $1.9 \pm (0.7)$ \\
&& IMTL-L   &f&a &&  \underline{0.744} &  \textbf{0.934} &&    0.0146 &      33.8 && avg &&   $6.5 \pm (0.3)$ \\
&& UW     &f&a &&  \textbf{0.746} &  \textbf{0.934} &&    0.0145 &      34.9 && avg &&   $7.2 \pm (1.9)$ \\
&& \underline{UW-SO}   &f&a &&  0.711 &  0.920 &&    \textbf{0.0127} &      \textbf{26.9} && avg &&  $\pmb{-1.4 \pm (0.6)}$ \\
  
\bottomrule
\end{tabular}

\label{tab:city_results_main}
\end{table}

\subsection{Common loss weighting methods benchmark}
\label{sec:comparison_common_lw_methods}
We compare our methods to the most common loss weighting approaches. 
Overall, UW-SO consistently performs best or second-best across all datasets and architectures w.r.t. the $\Delta_m$ metric. This holds for the hyperparameter-tuned experiments as well as for those with the fixed training protocol.
Occasionally, our method gets beaten by the computationally expensive Scalarization approach, especially when using the SegNet architecture.  
Noticeably, performance differences between MTO algorithms decrease for larger networks.

\begin{table*}[t]
\centering
\tiny 
\setlength\tabcolsep{2pt} 
\newcommand{\gcs}{\hspace{1pt}}
\caption{Test data results on NYUv2 with SegNet, ResNet-50, ResNet-101, and MTAN. Our method is underlined. For the first 3 architectures, we report the average over 5 runs, for $\Delta_m$ including $\pm$ one std. dev. Best score is in bold, second best underlined. LR and WD follow the schema as in Table \ref{tab:city_results_main}. The best softmax temperature for the 4 architectures is $T=3/2/3/2$, for Scalarization best weights for all 4 architectures are $\omega = [0.8, 0.1, 0.1]$ for segm., depth, and normal. Additionally, we report results over 3 seeds based on the training protocol that uses MTAN+SegNet with fixed hyperparameters \cite{liu2021conflict}.}
\begin{tabular}{cclll@{\gcs}ccc@{\gcs}ccc@{\gcs}cccccc@{\gcs}cc@{\gcs}cg}
\toprule
               \multirow{3}{*}{A.} & \multirow{3}{*}{E.} & \multirow{3}{*}{Method} & \multirow{3}{*}{lr} & \multirow{3}{*}{wd} && \multicolumn{2}{c}{Segmentation $\uparrow$} && \multicolumn{2}{c}{Depth $\downarrow$} && \multicolumn{5}{c}{Surface Normal} && \multirow{3}{*}{Ep} \\ 
                     &  & &  & && & && & &&  \multicolumn{2}{c}{Ang Dist $\downarrow$} & \multicolumn{3}{c}{Within $t^{\circ}$ $\uparrow$} &\\
                     & &  & & &&        mIoU & PixAcc &&        AbsErr & RelErr &&           Mean & Med &           11.25 &    22.5 & 30 && && $\Delta_m\%\downarrow$\\
\midrule
 \multirow{11}{*}{\rotatebox[origin=c]{90}{SegNet}} & \multirow{11}{*}{\rotatebox[origin=c]{90}{early stopping}} & STL  & e/f  & d &&         $0.361$ & $0.625$ &&        $0.601$ & $0.252$ &&         $24.8$ & $18.3$ &         $0.312$ & $0.585$ & $0.701$  \\ \hline

& & Scalar &   e &  f &&  \textbf{0.419} &  \textbf{0.675} &&     0.502 &     \underline{0.204} &&  \textbf{24.9} &    \textbf{19.0} &   \textbf{0.299} &  \textbf{0.573} &  \textbf{0.694} &&      360.4 && $\pmb{-5.3 \pm 0.7}$ \\
& & EW     &   e &   h &&  0.411 &  \underline{0.666} &&     0.512 &     0.207 &&  28.1 &    23.2 &   0.231 &  0.487 &  0.618 &&      380.0 &&   $4.6 \pm 2.5$ \\
& & RLW    &   e &  f &&  0.394 &  0.655 &&     0.523 &     0.208 &&  28.9 &    24.3 &   0.220 &  0.467 &  0.599 &&      388.8 &&   $7.7 \pm 2.3$ \\
& & DWA    &   e &   h &&  \underline{0.412} &  \underline{0.666} &&     0.509 &     0.207 &&  28.0 &    23.0 &   0.234 &  0.490 &  0.621 &&      376.4 &&   $4.2 \pm 2.1$ \\
& & GLS    &  f &  f &&  0.380 &  0.640 &&     0.516 &     0.211 &&  26.8 &    21.4 &   0.261 &  0.521 &  0.649 &&      372.0 &&  $ 2.4 \pm 1.8$ \\
& & IMTL-L &   e &   h &&  0.395 &  0.656 &&     \textbf{0.498} &     \underline{0.204} &&  26.4 &    20.8 &   0.269 &  0.534 &  0.660 &&      363.0 &&  $-0.3 \pm 1.5$ \\
& & UW     &   e &   h &&  0.394 &  0.655 &&     \underline{0.499} &     \textbf{0.201} &&  26.4 &    20.8 &   0.269 &  0.535 &  0.661 &&      386.0 &&  $-0.3 \pm 2.6$ \\
& & \underline{UW-SO}  &   e &  f &&  0.405 &  \underline{0.666} &&     0.508 &     0.205 &&  \underline{25.7} &    \underline{20.0} &   \underline{0.282} &  \underline{0.551} &  \underline{0.675} &&      363.8 &&  $\underline{-2.3 \pm 1.1}$ \\
\midrule
 \multirow{8}{*}{\rotatebox[origin=c]{90}{ResNet-50}} & \multirow{8}{*}{\rotatebox[origin=c]{90}{early stopping}}

& Scalar &   f &  a &&  0.480 &  0.717 &&     0.431 &     0.168 &&  \textbf{25.4} &    \underline{19.6} &   \textbf{0.298} &  \underline{0.557} &  \underline{0.677} &&      357.6 &&   $\underline{-9.7 \pm 0.2}$ \\
& & EW     &   e &   h &&  0.478 &  0.718 &&     0.428 &     \underline{0.166} &&  26.5 &    20.9 &   0.282 &  0.531 &  0.652 &&      334.2 &&   $-7.1 \pm 0.4$ \\
& & RLW    &   e &  f &&  \underline{0.482} &  \underline{0.720} &&     0.433 &     0.168 &&  26.6 &    21.1 &   0.279 &  0.527 &  0.649 &&      387.0 &&   $-6.6 \pm 0.3$ \\
& & DWA    &  f &   b &&  0.478 &  0.717 &&     0.431 &     0.168 &&  26.5 &    21.0 &   0.280 &  0.528 &  0.650 &&      329.2 &&   $-6.6 \pm 0.3$ \\
& & GLS    &   f &  f &&  0.469 &  0.711 &&     \underline{0.425} &     \underline{0.166} &&  \underline{25.8} &    19.9 &   \underline{0.296} &  0.550 &  0.668 &&      347.4 &&   $-8.7 \pm 0.4$ \\
& & IMTL-L &  f &   h &&  0.466 &  0.708 &&     0.434 &     \underline{0.166} &&  26.0 &    20.2 &   0.290 &  0.544 &  0.664 &&      362.0 &&   $-7.7 \pm 0.3$ \\
& & UW     &   e &   b &&  \textbf{0.485} &  \textbf{0.722} &&     0.433 &     0.169 &&  26.4 &    20.8 &   0.282 &  0.532 &  0.654 &&      324.4 &&   $-7.2 \pm 0.2$ \\
& & \underline{UW-SO}  &   f &  a &&  0.477 &  0.713 &&     \textbf{0.424} &     \textbf{0.165} &&  \textbf{25.4} &    \textbf{19.5} &   \textbf{0.298} &  \textbf{0.558} &  \textbf{0.678} &&      333.8 &&  $\pmb{-9.8 \pm 0.2}$ \\

\midrule
\multirow{8}{*}{\rotatebox[origin=c]{90}{ResNet-101}} & \multirow{8}{*}{\rotatebox[origin=c]{90}{early stopping}}

& Scalar &  e &   b &&  0.500 &  0.732 &&  0.417 &  0.159 &&  \textbf{24.9} &  \textbf{19.0} &  \textbf{0.306} &  \textbf{0.570} &  \textbf{0.689} && 341.8 &&   $\pmb{-12.5 \pm 0.3}$ \\
& & EW     &  e &  f &&  0.499 &  \textbf{0.734} &&     0.415 &     \underline{0.158} &&  25.9 &    20.1 &   0.291 &  0.546 &  0.666 &&      389.0 &&  $-10.1 \pm 0.2$ \\
& & RLW    &  e &     a &&  0.499 &  0.731 &&     0.415 &     \underline{0.158} &&  26.1 &    20.5 &   0.286 &  0.539 &  0.661 &&      372.6 &&   $-9.4 \pm 0.2$ \\
& & DWA    &  e &     a &&  0.498 &  \underline{0.733} &&     0.416 &     \underline{0.158} &&  25.9 &    20.2 &   0.290 &  0.544 &  0.665 &&      364.2 &&   $-9.8 \pm 0.3$ \\
& & GLS    &  e &  f &&  0.491 &  0.727 &&     \textbf{0.410} &     \textbf{0.157} &&  \underline{25.3} &    \underline{19.4} &   \underline{0.303} &  0.560 &  0.678 &&      329.8 &&  $-11.6 \pm 0.4$ \\
& & IMTL-L &  e &   b &&  \textbf{0.503} &  \underline{0.733} &&     0.415 &     \underline{0.158} &&  25.8 &    20.1 &   0.292 &  0.546 &  0.667 &&      306.0 &&  $-10.3 \pm 0.3$ \\
& & UW     &  e &   b &&  \underline{0.502} &  \underline{0.733} &&     0.414 &     \underline{0.158} &&  25.8 &    20.1 &   0.292 &  0.547 &  0.667 &&      325.2 &&  $-10.4 \pm 0.2$ \\
& & \underline{UW-SO}  &  e &   d && 0.494 &  0.725 &&     \underline{0.412} &     \textbf{0.157} &&  \textbf{24.9} &    \textbf{19.0} &   \textbf{0.306} &  \underline{0.569} &  \underline{0.687} &&      299.2 &&  $\underline{-12.3 \pm 0.3}$\\

\midrule
\multirow{8}{*}{\rotatebox[origin=c]{90}{MTAN/SegNet}} & \multirow{8}{*}{\rotatebox[origin=c]{90}{avg. last 10 ep.}}

&Scalar & f & a  &&  \textbf{0.400} &  \textbf{0.660} &&     \underline{0.534} &     \underline{0.225} &&   \underline{25.7} &     \underline{20.6} &   \underline{0.267} &  \underline{0.539} &  \underline{0.671} && avg &&  $\pmb{-1.5\pm 0.9}$ \\
& &EW   & f & a    &&  0.387 &  0.648 &&     0.575 &     0.247 &&   28.0 &     23.6 &   0.225 &  0.479 &  0.615 && avg &&  $7.0 \pm 1.2$ \\
& &RLW  & f & a    &&  0.382 &  0.637 &&     0.584 &     0.248 &&   28.5 &     24.3 &   0.216 &  0.465 &  0.602 && avg &&  $9.0 \pm 2.8$ \\
& &DWA  & f & a    &&  \underline{0.391} &  \underline{0.649} &&     0.586 &     0.250 &&   27.8 &     23.4 &   0.228 &  0.483 &  0.618 && avg &&  $6.8 \pm 0.7$ \\
& &GLS  & f & a    &&  0.384 &  \underline{0.649} &&     \textbf{0.528} &     \textbf{0.224} &&   26.7 &     21.9 &   0.249 &  0.511 &  0.644 && avg &&  $1.9 \pm 1.1$ \\
& &IMTL-L & f & a    &&  0.376 &  0.637 &&     0.572 &     0.248 &&   26.4 &     21.3 &   0.261 &  0.523 &  0.653 && avg &&  $2.7 \pm 1.3$ \\
& &UW   & f & a    &&  0.382 &  0.646 &&     0.550 &     0.236 &&   26.4 &     21.3 &   0.259 &  0.524 &  0.655 && avg &&  $1.6 \pm 1.3$ \\
& &\underline{UW-SO} & f & a    &&  0.351 &  0.626 &&     0.556 &     0.229 &&   \textbf{25.4} &     \textbf{19.8} &   \textbf{0.285} &  \textbf{0.555} &  \textbf{0.680} && avg && $\underline{-0.7 \pm 0.7}$ \\

\bottomrule
\end{tabular}

\label{tab:nyuv2_results_main}
\end{table*}
\begin{table}[t]
\centering
\scriptsize 
\setlength\tabcolsep{6pt} 
\caption{Test data results on CelebA with ResNet-18. 
We show the average test error (5 runs) over all 40 tasks. The chosen softmax temperature for UW-SO is $T=100$. We exclude GLS and Scalarization due to infeasibility, see details in the text.
}
\begin{tabular}{cclrrrrg}
\toprule
               \multirow{1}{*}{A.} & \multirow{1}{*}{E.} & \multirow{1}{*}{Method} & \multirow{1}{*}{lr} & \multirow{1}{*}{wd} & \multirow{1}{*}{Avg Err $\downarrow$} & \multirow{1}{*}{Ep} \multirow{1}{*} &{$\Delta_m\%\downarrow$} \\
 
\midrule
 \multirow{8}{*}{\rotatebox[origin=c]{90}{ResNet-18}} & \multirow{8}{*}{\rotatebox[origin=c]{90}{early stop.}} & STL  &g & f&        $9.24$ \\ \hline
 && EW     &    j &  f &           \underline{9.00} &       33.2 &  $-2.4 \pm 0.4$ \\
 &&RLW    &   h &  f &           9.01 &       54.4 &  $\underline{-2.5 \pm 0.6}$ \\
 &&DWA    &    j &  f &           9.01 &       39.2 &  $\underline{-2.5 \pm 0.7}$ \\
 &&IMTL-L &   h &     a &           9.18 &        5.0 &  $-1.2 \pm 0.9$ \\
 &&UW     &  g &   h &           9.26 &        6.0 &  $-0.1 \pm 0.7$ \\
 &&\underline{UW-SO}  &  g &  f &           \textbf{8.95} &       61.2 &  $\pmb{-4.0 \pm 0.2}$ \\
\bottomrule
\end{tabular}

\label{tab:celebA_results_main}
\end{table}

\textbf{Cityscapes}
On Cityscapes, UW-SO achieves the best $\Delta_m$ score when trained on both ResNet architectures as well as on the MTAN/SegNet with the fixed hyperparameters, and the second-best behind Scalarization when trained on the SegNet (see Table \ref{tab:city_results_main}).
In contrast to NYUv2 (see Table \ref{tab:nyuv2_results_main}), the performance of Scalarization decreases for larger networks due to weak results on the difficult and highly sensitive relative depth error. 
While an even more fine-grained search of task weights might yield better results, we argue that our weight search with step size going down to $0.02$ was performed adequately well to keep the computational cost feasible. 

\textbf{NYUv2}
We observe comparable results on NYUv2 in Table \ref{tab:nyuv2_results_main}, where UW-SO is again always best or second-best behind Scalarization. 
Both methods perform particularly well on the normal task, while still achieving strong results on the other two tasks. The fixed training protocol with MTAN (see last block in Table~\ref{tab:nyuv2_results_main}) leads to a slightly different order of goodness, especially IMTL-L ranks lower compared to the SegNet results.

\textbf{CelebA}
Considering a more challenging setup with 40 tasks, UW-SO is clearly exceeding the performance of all other methods with a $\Delta_m$ score of $-4.0$ and an average error of $8.95$ (see Table \ref{tab:celebA_results_main}). 
In contrast to the other two datasets, we did not include Scalarization due to the infeasibility of performing a grid search over 40 task weights. While we tried to run 50 different random weight combinations, we were not able to beat the EW performance and thus omit to report these results.
Furthermore, GLS is also not reported as the losses diverge due to numerical instabilities for a large number of tasks. 
In contrast to our previous results on other datasets, EW, RLW, and DWA show strong performance compared to UW and IMTL-L. 
We attribute this observation to UW and IMTL-L being prone to overfitting on some tasks,
as indicated by how early the validation $\Delta_m$ score reaches its minimum (e.g., epoch 6 for UW). 
We analyze the overfitting behavior of UW in comparison to UW-SO later in this section. 
As indicated by the negative $\Delta_m$ score, one achieves a positive transfer by training on multiple tasks simultaneously. Examining the task-level performance to verify that the enhanced average performance of UW-SO is not solely attributable to a limited set of tasks, it is noteworthy that UW-SO surpasses UW/IL/IMTL-L/RLW/EW/DWA in 34/34/31/27/24/24 out of 40 tasks.

\begin{figure}[t]
     \centering
     \begin{subfigure}[t]{0.35\textwidth}
         \centering
         \includegraphics[trim = 0mm 0mm 0mm 0mm, clip, width=\textwidth]{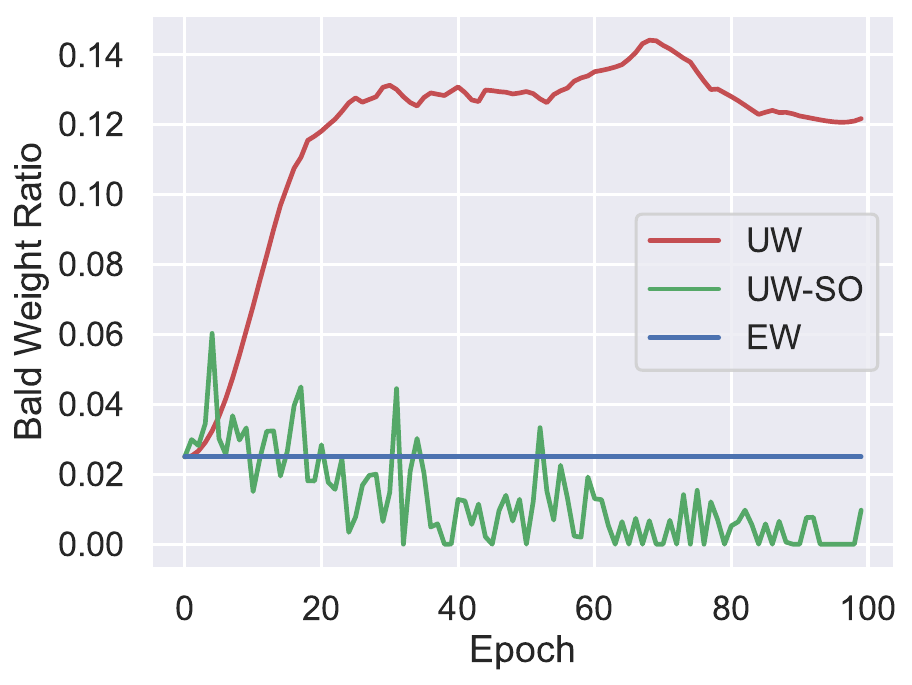}
         \caption{Weight ratio}
         \label{fig:celebA_bald_weight}
     \end{subfigure}
     \hfill
     \begin{subfigure}[t]{0.35\textwidth}
         \centering
         \includegraphics[trim = 0mm 0mm 0mm 0mm, clip, width=\textwidth]{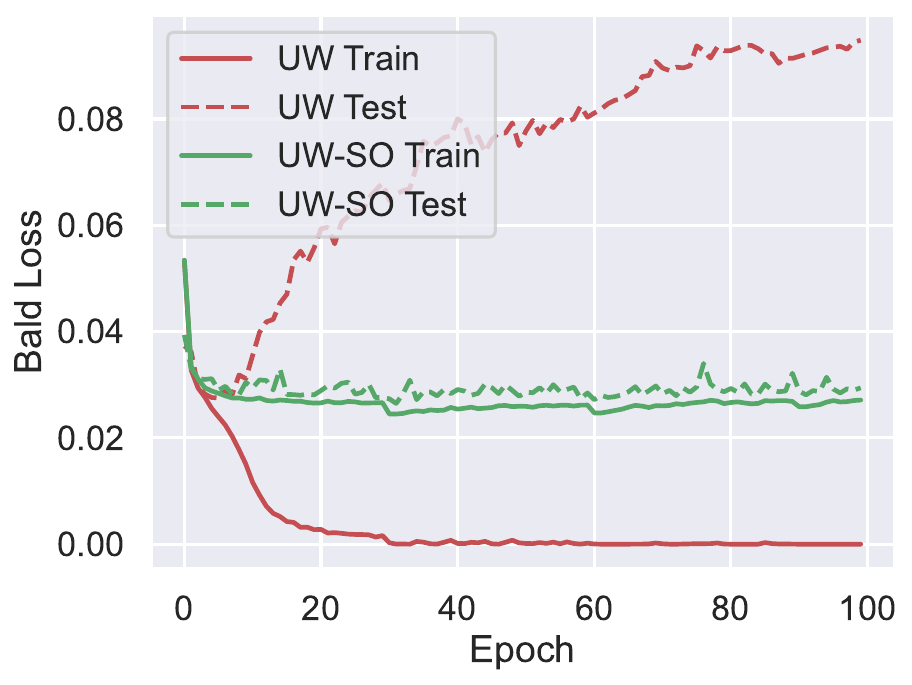}
         \caption{Loss}
     \end{subfigure}
     \hfill
    \caption{Comparison of weight ratio and loss development of UW and UW-SO for the \textit{Bald} task of CelebA. While UW shows superior training performance caused by putting a high weight on the task, it fails to generalize to unseen data (increasing test loss). 
    UW-SO puts less weight on the task and alleviates the overfitting.
    }
    \label{fig:celebA_bald_task}
\end{figure}

\textbf{Overfitting of UW}
\label{sec:overfitting}
Following our results in Table~\ref{tab:celebA_results_main}, UW achieves the worst $\Delta_m$ score for CelebA. We investigate the reasons: 
Figure~\ref{fig:celebA_bald_task} shows the train and test loss as well as the weight ratio of the \emph{Bald} task for UW and UW-SO. UW is subject to strong overfitting, as indicated by the huge gap between train and test loss. 
Contrary, UW-SO steadily decreases its training loss on the bald task, achieving its best test loss of $0.026$ at epoch 31 whereas UW has its best test loss of $0.028$ at epoch 5. For CelebA, for 34 out of 40 tasks, UW-SO achieves a lower test loss than UW and we assume this is due to overfitting - for most tasks, the train loss for UW drops to nearly $0$ (see Figure~\ref{fig:appx_celebA_losses} for all 40 tasks). 
Related to this, we observe that UW distributes much of the relative weight to only a few tasks, as can be seen for the \textit{Bald} task in Figure~\ref{fig:celebA_bald_weight}. All tasks are shown in~Figure~\ref{fig:appx_celebA_weights} and demonstrate a similar behaviour.

\textbf{Stronger networks and MTO approaches}
In Tables~\ref{tab:city_results_main} and \ref{tab:nyuv2_results_main}, 
our empirical analysis reveals an interesting trend: as network architectures increase in capacity, the influence of the MTL weighting method diminishes,
e.g., the difference between best and worst $\Delta_m$ score on NYUv2 is $13.0$ on the SegNet, but only $3.4 / 3.1$ on ResNet-50 / ResNet-101. 
This finding raises the question about the necessity of loss weighting methods for networks with large capacity.  
Further research in this direction is necessary.

\subsection{Ablation Studies}
\label{sec:ablation_studies}

In the following, we present some ablation studies about our method UW-SO. Further ablation studies can be found in sections \ref{sec:appx_task_weights} till \ref{sec:appx_performance_seeds}.

\textbf{Influence of softmax}
To demonstrate the influence of the softmax function on the inverse loss weights, we compare the $\Delta_m$ scores using UW-O and UW-SO across all datasets and architectures in Table \ref{tab:IL_vs_STIL}. UW-SO outperforms UW-O in all experiments, indicating the performance gain provided by the tempered softmax function. However, we want to emphasize that this is not due to a significantly worse performance of UW-O compared to other MTO methods. Therefore, we also present the results using IMTL-L, which, like UW-O, aims to scale each weighted task loss to 1, but unlike UW-O, it learns rather than computing the weights.
Interestingly, none of the two methods can outperform the other one, indicating that despite UW-O's simplicity, it still provides reasonable results compared to existing methods. Furthermore, in a toy example, we find that UW-O is particularly strong in dealing with extreme loss magnitude differences, at which most of the previous methods fail (Section~\ref{appx:motivation_uwo}). However, this is of less practical importance as losses usually show fewer differences in magnitudes due to normalized inputs than in our toy example.

\begin{table}[t]
\centering
\scriptsize 
\setlength\tabcolsep{6pt} 
\caption{Comparison of test $\Delta_m$ scores of UW-O, UW-SO, and IMTL-L across all evaluated datasets.}
\begin{tabular}{lrrrrrrr}
\toprule
\multirow{2}{*}{Method} & \multicolumn{3}{c}{\textbf{NYUv2}} & \multicolumn{3}{c}{\textbf{Cityscapes}} & \textbf{CelebA} \\
\cmidrule(lr){2-4} \cmidrule(lr){5-7} \cmidrule(lr){8-8}
& SN & RN-50 & RN-101 & SN & RN-50 & RN-101 & RN-18 \\
\midrule
IMTL-L & \underline{-0.3} & -7.7 & -10.3 & 10.6 & \underline{0.7} & \underline{1.6} & \underline{-1.2} \\
UW-O & 0.0 & \underline{-9.1} & \underline{-11.9} & \underline{5.5} & 2.1 & 2.8 & -0.6 \\
UW-SO & \textbf{-2.3} & \textbf{-9.8} &  \textbf{-12.3} & \textbf{4.4}  & \textbf{0.3} & \textbf{1.1} & \textbf{-4.0} \\
\bottomrule
\end{tabular}
\label{tab:IL_vs_STIL}
\end{table}

\begin{figure}[t]
     \centering
     \begin{subfigure}[t]{0.35\textwidth}
         \centering
         \includegraphics[trim = 0mm 0mm 0mm 0mm, clip, width=\textwidth]{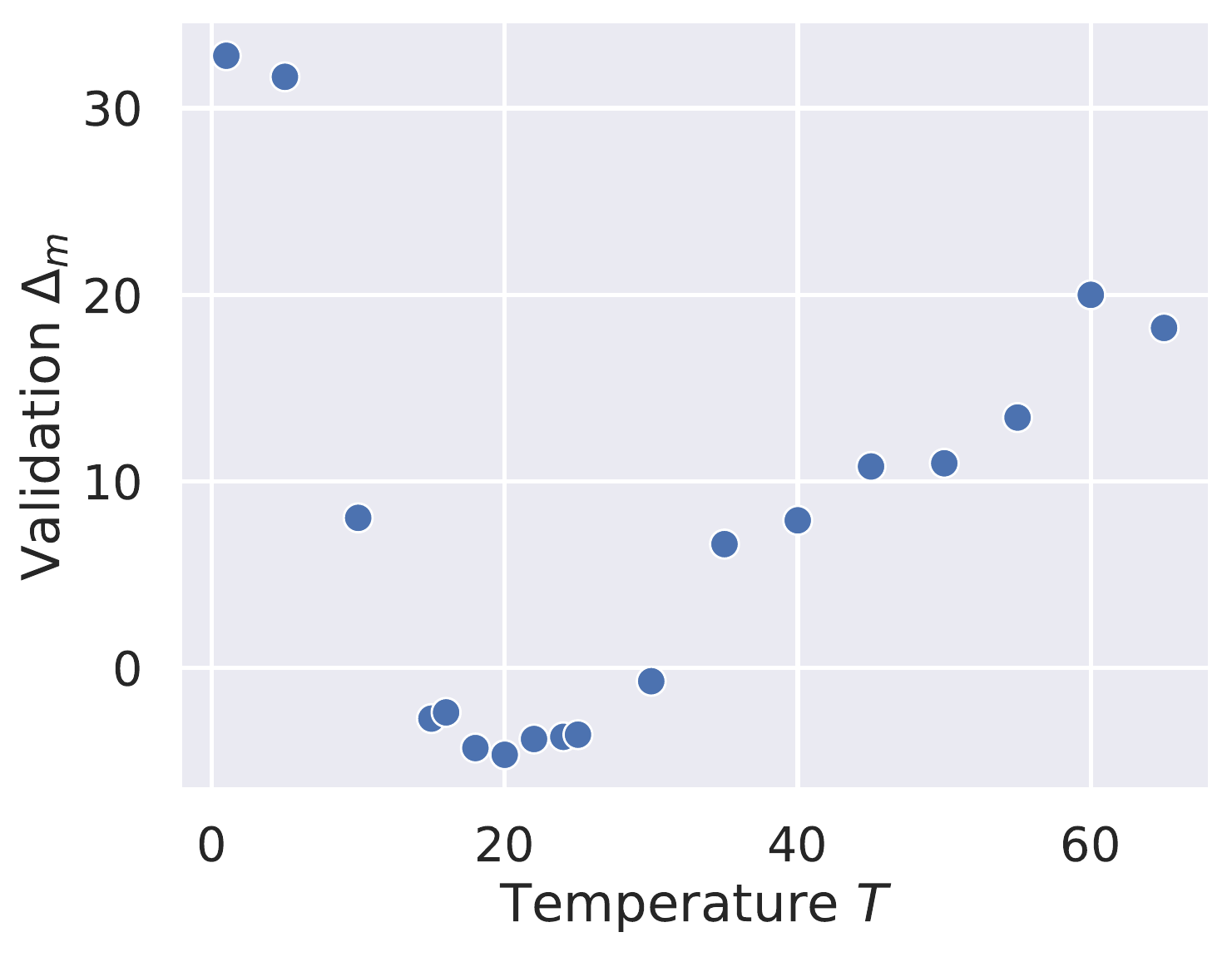}
         \caption{Cityscapes SegNet}
         \label{fig:abl_T_a}
     \end{subfigure}
     \hfill
     \begin{subfigure}[t]{0.35\textwidth}
         \centering
         \includegraphics[trim = 0mm 0mm 0mm 0mm, clip, width=\textwidth]{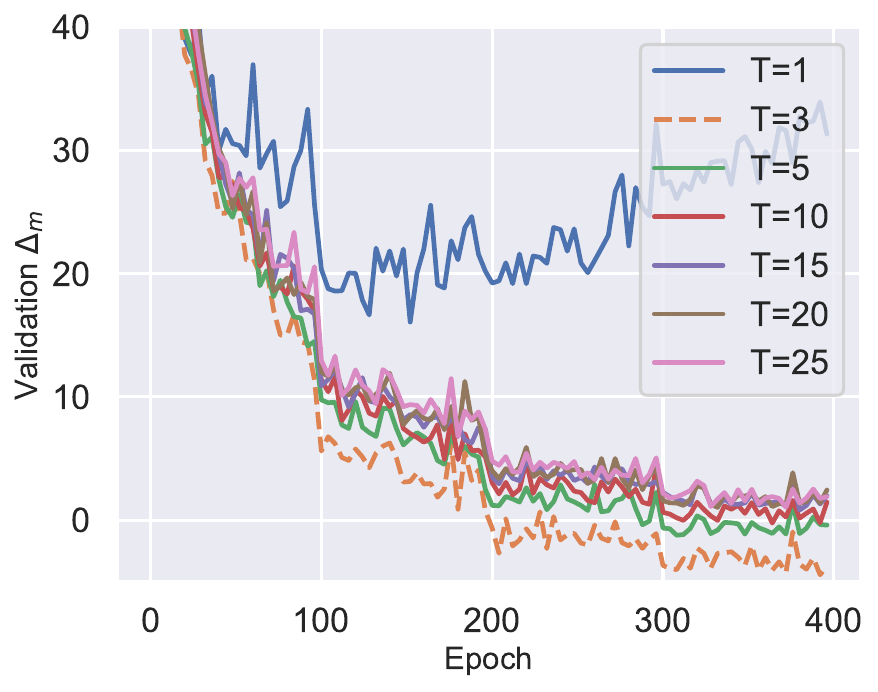}
         \caption{NYUv2 SegNet}
         \label{fig:abl_T_b}
     \end{subfigure}
     \hfill
    \caption{Performance of UW-SO for different choices of \textit{T} on the validation data. a) shows a clear, reasonably flat minimum for Cityscapes that eases the optimization of $T$. b) shows the $\Delta_m$ development for different \textit{T} values for NYUv2, indicating the optimal configuration already after around 100 epochs.
    }
    \label{fig:abl_T}
\end{figure}

\textbf{Influence of softmax temperature $\mathbf{T}$}
Our approach UW-SO has a strong performance, but one needs to tune the single hyperparameter $T$. Learning $T$ via gradient descent leads to suboptimal solutions in initial experiments. In Figure~\ref{fig:abl_T_a}, we show how the performance changes when the temperature $T$ is tuned for Cityscapes with a SegNet. The best run was achieved for $T=20$, but values close to it are also performing well. We conducted a line search for $T$ in steps of $5$ and further with a step size of $2$ around the optimum. We conclude that it is possible with acceptable tuning effort to find a good value for $T$. In Figure~\ref{fig:appx_abl_T_nyu} a similar behavior is shown for the three architectures on the NYUv2 dataset.

\textbf{Development of validation metric for different $\mathbf{T}$}
During analyses of the data, we observed that tuning of the hyperparameter $T$ can be eased by the following finding: Figure~\ref{fig:abl_T_b} shows the validation $\Delta_m$ score for different $T$ values over all epochs. Non-optimal $T$ values are clearly identifiable after around 100 epochs as having constantly a higher $\Delta_m$ score compared to favorable $T$ values. 
For instance, in this setup, it is reasonable to stop the runs for all initial $T$ values with step size of $5$ except for $T=5$ and $T=10$, and then proceed with a finer search around the optimum of which $T=3$ is best.
This reduces computational resources by a large amount. 
We additionally show results for ResNet-50 and ResNet-101 in Figure~\ref{fig:appx_searchT_rough} and~\ref{fig:appx_searchT_fine}. 
This clear separation of the best $T$ value cannot be made via the validation loss, as shown in Figure~\ref{fig:appx_searchT_rough_loss}.

\begin{figure}[t]
     \centering
     \begin{subfigure}[t]{0.3\textwidth}
         \centering
         \includegraphics[trim = 0mm 0mm 0mm 0mm, clip, width=\textwidth]{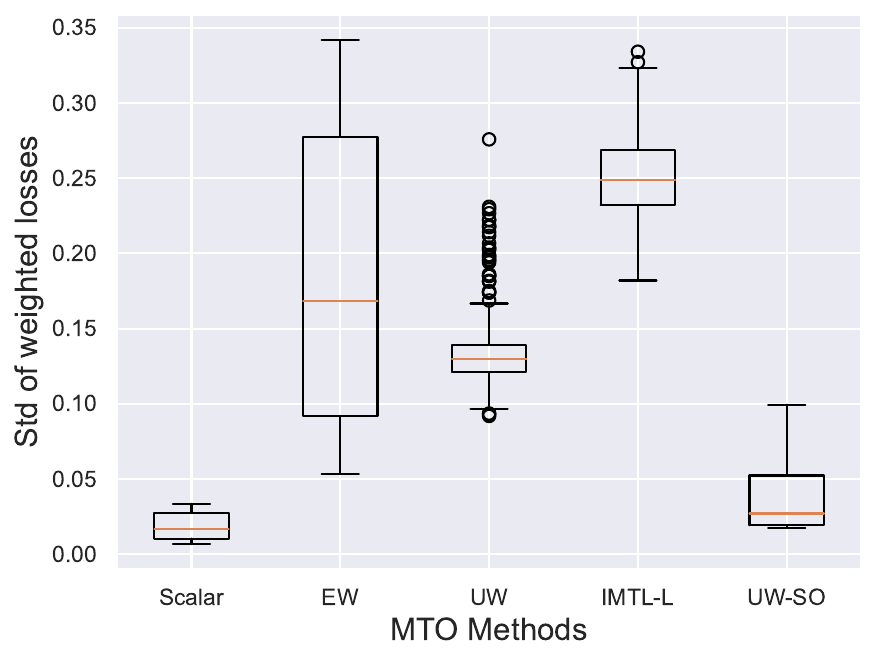}
         \caption{Segmentation}
     \end{subfigure}
     \hfill
     \begin{subfigure}[t]{0.3\textwidth}
         \centering
         \includegraphics[trim = 0mm 0mm 0mm 0mm, clip, width=\textwidth]{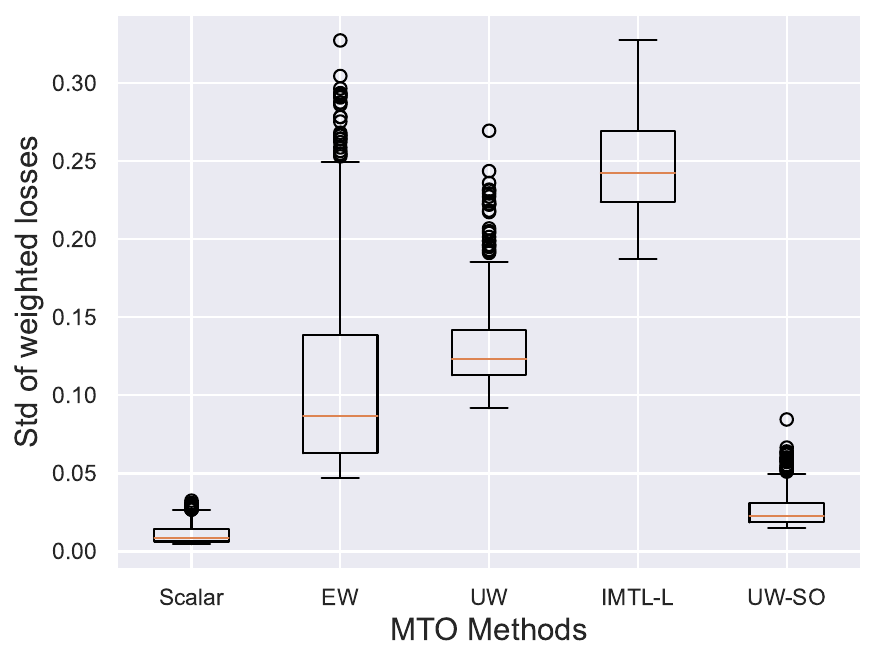}
         \caption{Depth}
     \end{subfigure}
     \hfill
    \begin{subfigure}[t]{0.3\textwidth}
         \centering
         \includegraphics[trim = 0mm 0mm 0mm 0mm, clip, width=\textwidth]{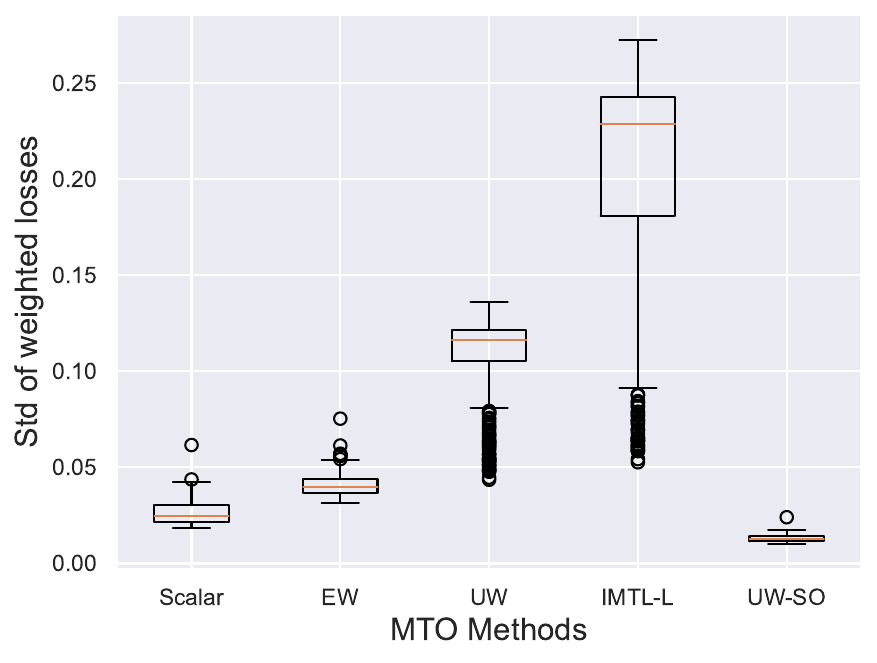}
         \caption{Normal}
     \end{subfigure}
     \hfill
    \caption{Boxplots over the std. dev. of the weighted NYUv2 (SegNet) task losses $\omega_k L_{k}$ of all batches from one epoch. Std. dev. over one epoch is one observation.}
    \label{fig:nyu_boxplots}
\end{figure}

\textbf{Oscillation of MTO methods} 
The authors of the IMTL-L approach argue that weighting by the inverse of the loss results in "severe oscillations" \cite[sec. 3.2]{liu2021towards}. Our experiments confirm that the gradient-based methods IMTL-L and UW have smoother loss weight updates than UW-O and UW-SO (see Figure~\ref{fig:appx_nyu_weights}). However, we argue that oscillations of the task weight $\omega_k$ itself are not problematic as it is the weighted loss $\omega_k L_k$, which determines the parameter update. Looking at the standard deviation over weighted losses from all batches within one epoch (Figure~\ref{fig:nyu_boxplots}), it turns out that UW-SO, and Scalarization are less affected by oscillations than IMTL-L (UW-O has a standard deviation of 0 by design, see Fig.~\ref{fig:appx_nyu_weighted_loss}).

\section{Discussion and Conclusion}
\label{sec:exp_conclusion}
In this paper, we introduce UW-SO, a new method for weighting losses in Multi-Task Learning (MTL). Derived from the analytical solution of Uncertainty Weighting, UW-SO applies the tempered softmax function to the inverse of the losses to effectively weight tasks. In an extensive benchmark with 3 datasets, up to 4 architectures per dataset, and 8 different loss weighting methods (focussing on pure loss-weighting and not gradient-based methods), we demonstrate that UW-SO achieves superior results. Only the brute-force Scalarization approach could occasionally challenge UW-SO, though, Scalarization is not feasible in practice due to the immense tuning demand for a large number of tasks. 

Furthermore, our evaluation reveals insights into the training behavior of existing weighting methods, indicating that larger networks lead to less pronounced differences among MTL methods, and that learning rate tuning for each weighting approach is essential while weight decay tuning seems less influential.

We hope that our benchmark lays the ground for fruitful future discussion on MTL and gives guidance for practitioners. 
Future investigations should focus on further reducing the computational demands of weighting methods while preserving performance, e.g., heuristics to determine a good value for $T$ in UW-SO presents a promising avenue for research. Furthermore, it remains open whether there is a threshold for the "strength" of a network at which MTL weighting methods no longer significantly influence performance.

%
%
%
%
\bibliographystyle{splncs04}
\bibliography{main}







\clearpage
\newpage
\onecolumn

\setlength{\belowcaptionskip}{0pt} 

\begin{center}
    \large  
    \textbf{Supplementary Material} 
\end{center}

\renewcommand{\thetable}{A\arabic{table}}
\renewcommand{\thefigure}{A\arabic{figure}}
\renewcommand{\thesection}{A\arabic{section}}

\setcounter{table}{0}
\setcounter{figure}{0}
\setcounter{section}{0}
\setcounter{page}{1} 

\section{Derivation of UW-SO}
\label{sec:appx_derivation_uwso}

In this section, we provide more details on the analytically optimal derivation of the uncertainty-based task weights. Note that the analytical solution to UW, which we call UW-O, varies (just as UW) for different loss criteria (e.g., $L_2$ and Cross-Entropy Loss yield a different constant in the denominator compared to $L_1$). However, we simplify this by taking a unified formula (see Eq. \ref{formula:UW-O}) not depending on the kind of task as also done in \cite{liu2021conflict,LibMTL} for UW. This reduces the implementation overhead. Empirically, we further saw small improvements in the results in first experiments due to the unification. 

Nevertheless, we show the detailed derivation of UW-O for $L_1$, $L_2$, and Cross Entropy Loss in the following paragraphs.

\paragraph{L1 loss}
In the case of regression tasks evaluated by the $L_1$ loss we define our likelihood as a Laplace distribution, thus, the objective is given as \cite{kendall2018multi}


\begin{equation}
    \min_{\sigma_k} \frac{1}{\sigma_k} L_k + \log \sigma_k,
\end{equation}
where we minimize the UW loss function with respect to $\sigma_k$.
Taking the derivative and solving for $\sigma_k$ results in an analytically optimal solution:

\begin{equation}
    \frac{\partial}{\partial \sigma_k} \frac{1}{\sigma_k} L_k + \log \sigma_k = -\frac{1}{\sigma_k^2}L_k + \frac{1}{\sigma_k}
\end{equation}

\begin{equation}
    -\frac{1}{\sigma_k^2}L_k + \frac{1}{\sigma_k} \overset{!}{=} 0
\end{equation}

\begin{equation}
    \sigma_k = L_k
\end{equation}

We assume $\sigma_k$ to be positive and therefore only allow for positive losses. Note that this limitation comes from UW which only works for losses that are positive and are based on a location scale distribution.

Replacing $\sigma_k$ with its analytical solution $L_k$ in the total loss function gives the following loss:

\begin{equation}
    L = \frac{1}{sg[L]} L + \log sg[L],
\end{equation}

where we denote $sg$ as the stopgradient operator to avoid zero gradients of the network updates. Since we do not compute any gradient of the second part of the loss, we can simplify the term, such that

\begin{equation}
    L = \frac{1}{sg[L]} L.
\end{equation}

To show that UW-O is indeed unaffected of loss scalings, we can assume task-specific weights $\omega_k$ in the loss function. However, these weights cancel out, making UW-O very effective for scenarios with highly imbalanced losses, such as in our experiment in Section \ref{appx:motivation_uwo}.

\begin{equation}
    L = \sum_{k \in \mathcal{K}} \frac{1}{\omega_k sg[L_k]} \omega_k L_k = \sum_{k \in \mathcal{K}} \frac{1}{sg[L_k]} L_k
    \label{eq:uw_o_invariance}
\end{equation}

\paragraph{L2 loss}
The objective in UW for regression tasks with an $L_2$ loss can be formulated as \cite{kendall2018multi}
\begin{equation}
    \min_{\sigma_k} \frac{1}{2\sigma_k^2} L_k + \log \sigma_k,
\end{equation}
Consequently, our derivation and total loss again change:
\begin{equation}
    \frac{\partial}{\partial \sigma_k} \frac{1}{2\sigma_k^2} L_k + \log \sigma_k = -\frac{1}{\sigma_k^3}L_k + \frac{1}{\sigma_k}
\end{equation}

\begin{equation}
    -\frac{1}{\sigma_k^3}L_k + \frac{1}{\sigma_k} \overset{!}{=} 0
\end{equation}

\begin{equation}
    \sigma_k = \pm \sqrt{L_k}
\end{equation}

\begin{equation}
    L = \frac{1}{2 sg[L]} L + \log \sqrt{sg[L]}
\end{equation}

\begin{equation}
    L = \frac{1}{2 sg[L]} L.
\end{equation}

\paragraph{Cross-Entropy Loss}
The derivation for classification tasks evaluated by the cross-entropy loss looks as follows:

\begin{equation}
    \min_{\sigma_k} \frac{1}{\sigma_k^2} L_k + \log \sigma_k
\end{equation}

\begin{equation}
    \frac{\partial}{\partial \sigma_k} \frac{1}{\sigma_k^2} L_k + \log \sigma_k = -\frac{2}{\sigma_k^3}L_k + \frac{1}{\sigma_k}
\end{equation}

\begin{equation}
    -\frac{2}{\sigma_k^3}L_k + \frac{1}{\sigma_k} \overset{!}{=} 0
\end{equation}

\begin{equation}
    \sigma_k = \pm \sqrt{2L_k}
\end{equation}

\begin{equation}
    L = \frac{1}{2 sg[L]} L
\end{equation}

\section{Implementation  details}
\label{sec:appx_impl_details}

\subsection{Data and optimization during training}

\textbf{NYUv2} contains 464 indoor scenes recorded in three different cities, resulting in 636 images for training, 159 for validation, and 654 images for testing. The dataset comprises three tasks: 13-class semantic segmentation, depth estimation, and surface normal prediction.
For training, we follow previous work \cite{liu2021conflict,liu2019end} and resize the images to 288x384 and apply image augmentations. The augmentations include randomly scaling the images (ratio=1.0, 1.2, 1.5) and randomly flipping them (p=0.5). We train all models on NYUv2 for 400 epochs (except for the MTAN/SegNet model, which follows the previous training protocol as in \cite{liu2021conflict} with 200 epochs, batch size of 2, Adam optimizer, and a learning rate of $10^{-4}$ which is halved after 100 epochs) with a batch size of 2 and apply the Adam optimizer with a specifically tuned learning rate $\gamma \in [10^{-3}, 5*10^{-4}, 10^{-4}, 5*10^{-5}, 10^{-5}]$ and weight decay $\lambda \in [0, 10^{-6}, 10^{-5}, 10^{-4}, 10^{-3}, 10^{-2}]$ for each weighting method. Similar to Liu et al. \cite{liu2021conflict}, we decay the learning rate after 100 epochs by a factor of 2.

\textbf{Cityscapes} consists of urban street scenes from 50 different cities yielding 2380 images for training, 595 for validation, and 500 images for testing. The dataset has 2 tasks: 7-class semantic segmentation and depth estimation. 
We resize the images to 128x256 and apply the same augmentation techniques used for NYUv2.
Models are trained for 600 epochs (except for the MTAN/SegNet model with the same specifications as detailed for NYUv2, just the batch size for Cityscapes is 8) with a batch size of 8 using the same optimizer and learning rate as well as weight decay grid as for NYUv2. Again, we halve the learning rate every 100 epochs. We follow Xin et al. \cite{xin2022current} and randomly sample 595 images from the training split as validation data and report test results on the original validation split.

\textbf{CelebA} is a dataset of celebrity faces with different attributes of 10,177 identities. We cast it as an MTL dataset by viewing the available 40 binary attributes, e.g., glasses and smiling as individual classification tasks. The train, validation, and test set contain 162,770, 19,867, and 19,962 samples, respectively.
Training on CelebA is performed for 100 epochs with a batch size of 256, use the Adam optimizer with tuned learning rate $\gamma \in [5*10^{-2}, 10^{-2}, 5*10^{-3}, 10^{-3}, 5*10^{-4}, 10^{-4}]$ and weight decay $\lambda \in [0, 10^{-5}, 10^{-4}, 10^{-3}]$. We halve the learning rate every 30 epochs.

\subsection{Architectures} The code for the SegNet architecture is based on the repo\footnote{ \url{https://github.com/Cranial-XIX/CAGrad/blob/main/cityscapes/model_segnet_split.py}} from the CAGrad method \cite{liu2021conflict}. The ResNet-18 implementation is based on the repo\footnote{\url{https://github.com/isl-org/MultiObjectiveOptimization/blob/master/multi_task/models/multi_faces_resnet.py}} from \cite{sener2018multi} and we base our code on a DeepLabV3+ implementation for the ResNet-50 architecture\footnote{\url{https://github.com/VainF/DeepLabV3Plus-Pytorch/blob/master/network/backbone/resnet.py}} in combination with the DeepLabV3+ architecture\footnote{ \url{https://github.com/VainF/DeepLabV3Plus-Pytorch/blob/master/network/_deeplab.py}}.

The heads of the architectures are as follows: Segnet \cite{badrinarayanan2017segnet} has task-specific last heads, i.e., for Cityscapes one convolutional layer for the segmentation and depth task respectively, and for NYUv2 one convolutional layer for each of the tasks (segmentation, depth, surface normal) as well. The ResNet-50 and ResNet-101 for NYUv2 use a DeepLabHead for each of the tasks. Each head of CelebA attached to the ResNet-18 is a linear layer.

ResNet-50 and ResNet-101 are pre-trained on ImageNet \cite{deng2009imagenet}. Other models are randomly initialized with the standard PyTorch initialization. We use version \textit{1.13.1+cu117}. 

The code for the MTAN on top of the SegNet is based on the LibMTL library \cite{LibMTL}.

\subsection{Hyperparameter optimization}
\begin{figure*}[ht!]
    \centering
    \begin{subfigure}{.31\textwidth}
        \includegraphics[width=\textwidth,keepaspectratio]{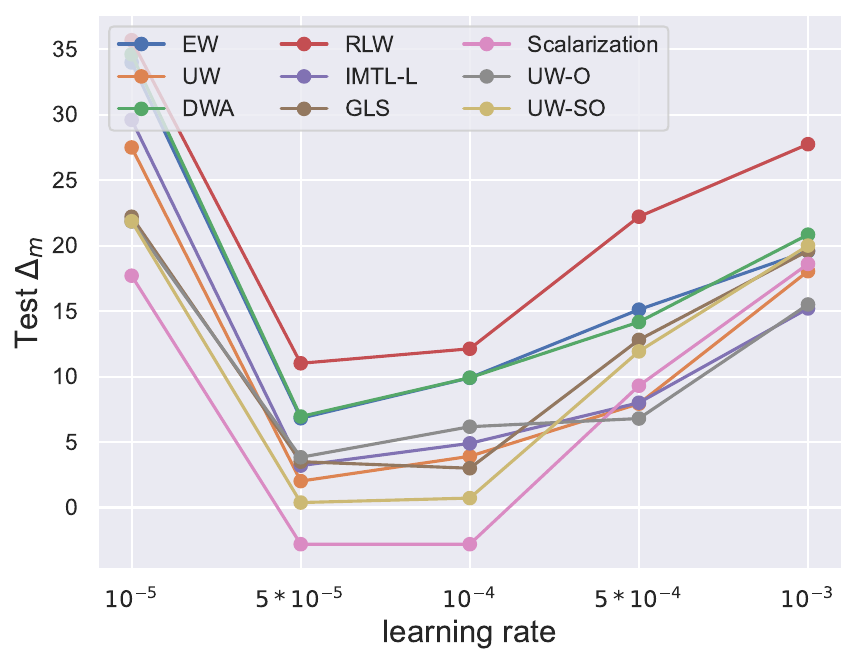}  
        \caption{NYUv2}
        \label{fig:nyu_lr}
    \end{subfigure}
    \begin{subfigure}{.31\textwidth}
        \includegraphics[width=\textwidth,keepaspectratio]{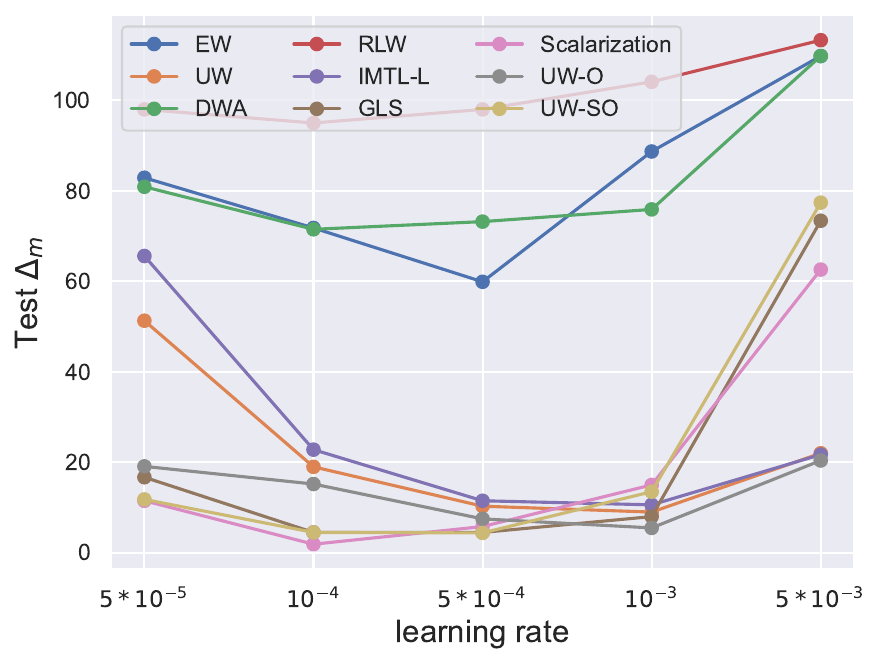}
        \caption{Cityscapes}
        \label{fig:city_lr}
    \end{subfigure}
    \begin{subfigure}{.31\textwidth}
        \includegraphics[width=\textwidth,keepaspectratio]{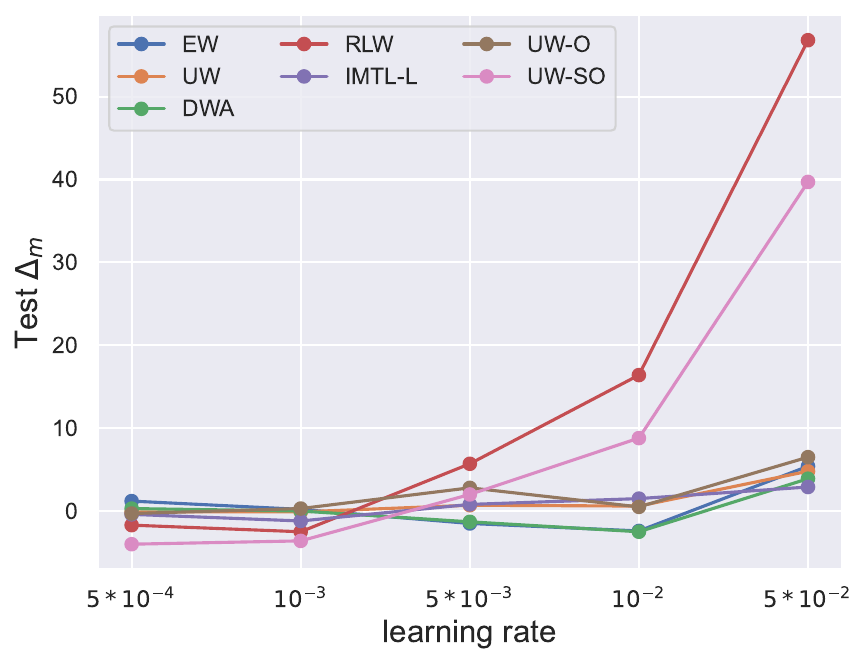}  
        \caption{CelebA}
        \label{fig:celebA_lr}
    \end{subfigure}
    \caption{$\Delta_m$ scores on the test data for different choices of the learning rate with a fixed weight decay (for (a) and (b): $\lambda = 10^{-5}$; for (c): $\lambda = 10^{-4}$) according to our chosen line search approach, averaged over 5 runs. We show results for a) NYUv2 with SegNet, b) Cityscapes with SegNet, and c) CelebA with ResNet-18. In particular for (b) and (c) the optimal learning rate value highly varies across different weighting approaches, underlining the necessity to perform method-specific learning rate tuning.}
  \label{fig:abl_lr}
\end{figure*}

\begin{figure}[ht]
    \centering
     \begin{subfigure}[t]{0.24\textwidth}
         \centering
         \includegraphics[trim = 0mm 0mm 0mm 0mm, clip, width=\textwidth]{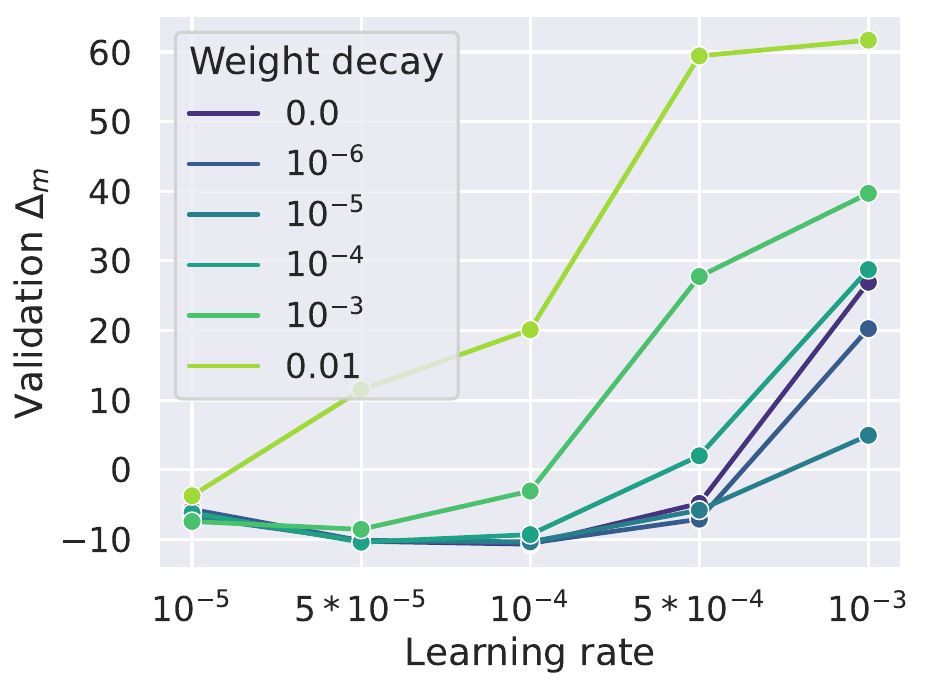}
         \caption{Scalarization}
     \end{subfigure}
     \hfill
    \begin{subfigure}[t]{0.24\textwidth}
         \centering
         \includegraphics[trim = 0mm 0mm 0mm 0mm, clip, width=\textwidth]{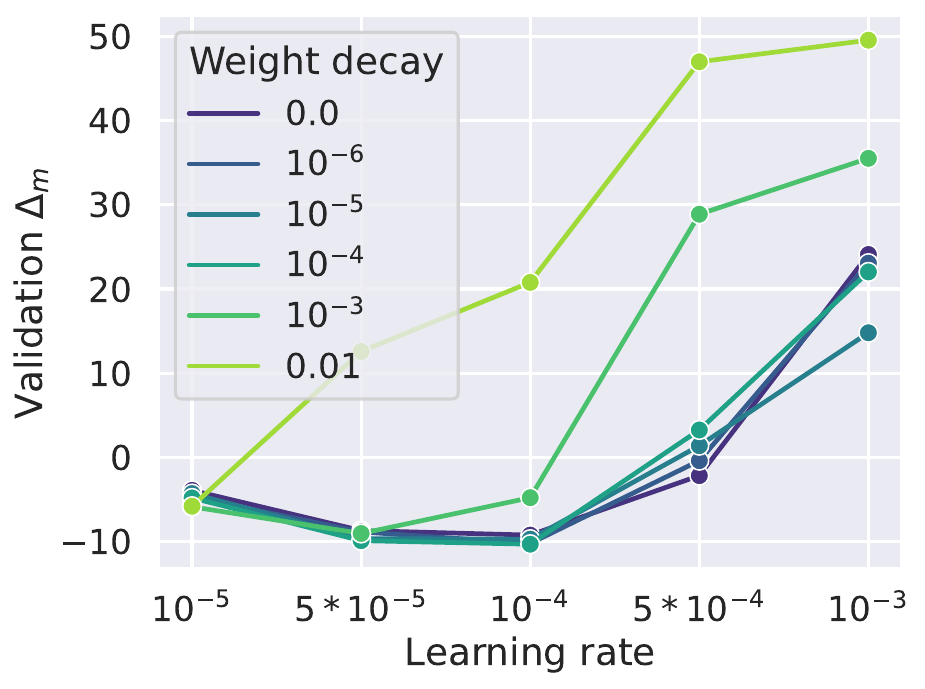}
         \caption{GLS}
     \end{subfigure}
     \hfill
    \begin{subfigure}[t]{0.24\textwidth}
         \centering
         \includegraphics[trim = 0mm 0mm 0mm 0mm, clip, width=\textwidth]{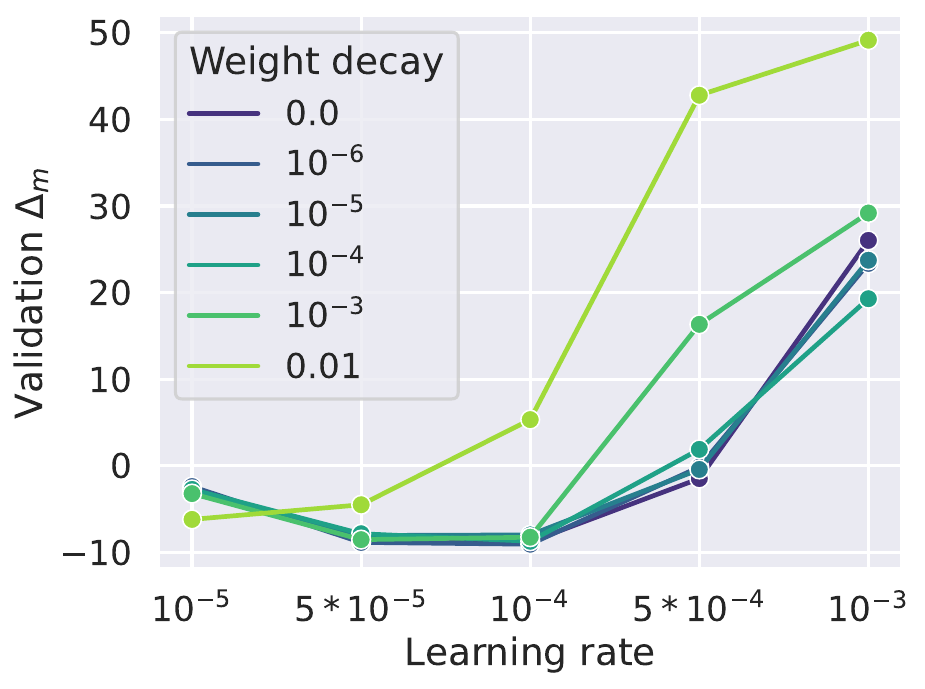}
         \caption{UW}
     \end{subfigure}
     \hfill
    \begin{subfigure}[t]{0.24\textwidth}
         \centering
         \includegraphics[trim = 0mm 0mm 0mm 0mm, clip, width=\textwidth]{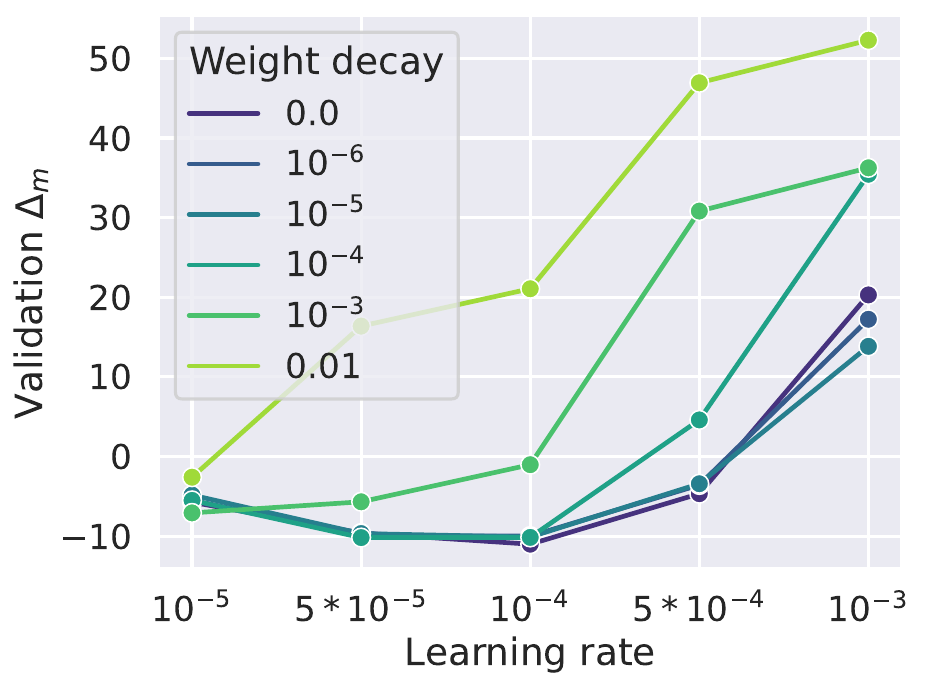}
         \caption{UW-SO}
     \end{subfigure}
     \hfill
    \caption{Results on NYUv2 with ResNet50 backbone when varying learning rate and weight decay.
    If the weight decay is chosen to be reasonably small (i.e. $\leq0.0001$), we observe only a low variation in performance for varying weight decays and a specific learning rate over all methods.
    Therefore, we argue it is sufficient to 
 perform line search one by one when tuning these hyperparameters.}
    \label{fig:appx_lr_wd}
\end{figure}

To avoid a huge computational overhead due to the large amount of hyperparameters involved in the MTO problem, we perform a line search to tune hyperparameters. From a practitioner's point of view, this is much more feasible than a grid search (as employed by Xin et al. \cite{xin2022current}) or other techniques. Thus, we perform an extensive and fair comparison of weighting methods in MTL and still keep computational costs as low as possible by using the following approach. First of all, we start with a fixed weight decay value for each combination of architecture and dataset. To get a reasonable initial weight decay value, we set the hyperparameter based on previous work. For example, we use $\lambda = 10^{-4}$ for all weighting methods trained with the ResNet-18 on the CelebA dataset, following Liu et al. \cite{liu2021towards}. Similarly, we set $\lambda = 10^{-5}$ for the NYUv2 and Cityscapes experiments, following Lin et al. \cite{lin2021reasonable}. Assuming the selected weight decay, we perform an extensive grid search of learning rate and weighting method-specific hyperparameters. Latter includes the scalars from the Scalarization approach and the softmax temperature $T$ from UW-SO. Doing so, we ensure to find the best combination of learning rate and task weights, which is crucial to achieving competitive performance (see Figure~\ref{fig:abl_lr} for the intermediate test results after tuning the learning rate and the method-specific hyperparameters with a fixed weight decay). After finding optimal values for learning rate $\gamma$ and task weights $\omega$, we finally tune the weight decay $\lambda$ using a line search. This ensures that each weighting method is equipped with the right amount of regularization to achieve the best possible results. Figure~\ref{fig:appx_lr_wd} empirically underlines that our line search approach does not suffer in performance compared to a grid search. This is because adapting the weight decay (if reasonably small, i.e., $\leq 0.0001$) results in only a very low variation in performance. In fact, for Scalarization, GLS, and UW-SO using the ResNet-50 we find the exactly same learning rate and weight decay with our line search approach compared to the more expensive grid search. Only for UW, we see a fairly small deviation as the line search results in a validation $\Delta_m$ score of -8.83 with $\gamma = 5*10^{-5}$, while the grid search finds a slightly better score of -9.03 with $\gamma = 10^{-4}$. However, we argue that this deviation is fairly small and even lies within the standard deviation caused by the random seeds. As a result, we are confident that our tuning process has been exhaustive and reliable.

In summary, the LR has a high impact on the performance while the influence of the WD is low in comparison. This underlines our line search approach to first tune the LR (with any method-hyperparameters, if present) and then tune the WD.

\subsection{Discussion on $\Delta_m$ metric}
Some MTO approaches improve the result of one task while lowering the performance of others. The $\Delta_m$ metric shows a balanced result of relative improvement among all tasks. It can thus be seen as an indication which MTO approach yields the best results if all single task metrics are equally important.
However, as some tasks are easier to improve on a relative scale, MTO methods focusing on those tasks achieve a better $\Delta_m$ score. 
Moreover, tasks with more metrics (e.g. the surface normal for NYUv2) have a higher influence on the $\Delta_m$ score, which is therefore biased towards these tasks (see Table~\ref{tab:nyuv2_results_main},  the best performing methods on the normal task also reach the highest $\Delta_m$ score). 

As $\Delta_m$ is currently the standard evaluation criteria and we have no specified prioritization of any task in our setting  we nevertheless report it as our main decision criteria.

\subsection{Two evaluation setups}
Several papers from 2018 to 2022 (e.g., \cite{liu2021conflict,liu2019end,yu2020gradient,navon2022multi}) have adopted a strict \textit{training protocol} for NYUv2 and Cityscapes to compare MTO methods, which trains with fixed hyperparameters (200 epochs, no tuning of LR, WD=0), without a validation set and report results as the average of the last 10 test epochs. 
On the contrary, another recent direction in the MTO domain (e.g., \cite{xin2022current,kurin2022defense,sener2018multi}) advocates strongly for method-specific hyperparameter tuning, which is of more relevance for the practitioner. 
Our decision to follow the latter group is based on the findings of \cite{xin2022current} that MTO methods are sensitive to LR and WD and not tuning those "can create a false perception of performance improvement" \cite[p.2]{xin2022current}. We confirm these results in Fig. \ref{fig:abl_lr} and \ref{fig:appx_lr_wd}.
Following \cite{xin2022current}, we perform a thorough hyperparameter search (LR and WD) for all methods choosing the best combination w.r.t. the best $\Delta_m$ score using early stopping on the validation set on one seed. 
For the final evaluation, we train on 5 random seeds and report mean test set performance, as done in \cite{xin2022current,kurin2022defense,navon2022multi}. However, acknowledging \cite{liu2021conflict,liu2019end,yu2020gradient,navon2022multi}, we also show results using their \textit{training protocol} with an MTAN/SegNet on NYUv2 and Cityscapes.

To the best of our knowledge, this is the first loss weighting benchmark that includes extensive method-specific LR and WD tuning. To give an intuition for the extensiveness, we evaluated 253 experiments only for NYUv2 with the SegNet (including the grid search of LR and MTO hyperparameters, WD search, and 5 seeds). However, acknowledging \cite{liu2021conflict,liu2019end,yu2020gradient,navon2022multi}, we also show results using their \textit{training protocol} with an MTAN/SegNet on NYUv2 and Cityscapes.

\section{Additional Results}
\subsection{Comparison of our results to existing publications}
\label{sec:comparison_other_publications}
In general, since there is no predefined training and validation protocol in MTL, it is difficult to compare results from different publications. 
Aspects such as the number of epochs, the backbone/head of the architecture (often also called encoder/decoder), the optimizer, additional regularizations, such as dropout layer, existing data augmentations or available values for the line search for LR and WD, influence the results, sometimes considerably, so that even publications on the same datasets can only be compared to a limited extent on the basis of their values. 

In~\cite{kurin2022defense} mainly gradient-based MTO approaches are compared. The only approach we can compare our results to is RLW with normal distribution. On the CelebA dataset, \cite{kurin2022defense} only train for 50 epochs on the ResNet-18, but also achieve an average error of about $9.01\%$ (\cite[Figure 2(a)] {kurin2022defense}). This fits well with our results as the early stopping epoch in RLW is on average $54.4$ (see Table \ref{tab:celebA_results_main}). Thus, we found the same result as \cite{kurin2022defense}, even though we performed a longer training. For Cityscapes, they report results using the DeepLabV3+ architecture with ResNet-50 as the backbone.
However, the reported results (\cite[Figure 3]{kurin2022defense}) are worse than our results for ResNet-50 in Table~\ref{tab:city_results_main}. This is likely due to the fact that they only train for 100 epochs, while we train for 600. However, their results are still not dramatically worse, although we see in Table \ref{tab:city_results_main} that the validation optimum is reached for Scalarization and RLW after 400 epochs, which is significantly higher than 100 epochs. We assume that this is because Kurin et al. \cite[p. 6]{kurin2022defense} "evaluate a different model for each metric, chosen as the one with the best (maximal or minimal, depending on the metric) validation performance across epochs (we perform per-run early stopping). This procedure maximizes per-task performance, at the cost of increased inference time. If inference time is a priority, an alternative model selection procedure could rely on relative task improvement, assuming that per-metric improvements are to be weighted linearly." This means each metric for Cityscapes is optimized on its own model, whereas we are taking one model (at the highest validation $\Delta_m$ epoch) for all metrics. Taking one separate model for each metric could overestimate the overall results, as we saw in our experiments that weighting single tasks higher (which is possible with Scalarization) can reduce the performance of other tasks, partly considerably.

The work of~\cite{navon2022multi} which introduces a new method termed \textit{Nash-MTL} would have results to compare on the NYUv2 and Cityscapes dataset for the MTO approaches RLW, UW, and DWA. However, their chosen settings (discussed in \cite[Section B, Appendix]{navon2022multi}) are considerably different from ours, such that a direct comparison is not possible. In \cite{navon2022multi} only 200 epochs are used for training on NYUv2 / Cityscapes compared to 400 / 600 we use. Additionally, the work uses a fixed learning rate of $10^{-4}$ for all methods, a practice which was shown in~\cite{xin2022current} to possibly bias comparisons of different methods. Moreover, an MTAN architecture is used which is not directly comparable to our backbone/head architectures. 

Xin et al. \cite{xin2022current} underline the strength of the Scalarization approach. As they also use early stopping on the validation data set to find the best model, methods comparable from their publication are RLW and Scalarization. For Cityscapes, they also use the DeepLabV3+ architecture with the ResNet-50 as backbone, detailed in ~\cite{lin2022rlw} which is also one of our architectures. Though, in \cite{xin2022current} the number of epochs is not stated explicitly, so we assume they use 200 epochs as in \cite{lin2022rlw}. Comparing the presented metrics for Cityscapes (see Figure 8 and Figure 15 in \cite{xin2022current}), they report values for mIoU / PxAcc / AbsErr using RLW at approximately: 0.695 / 0.917 / 0.0135, which we outperform probably due to a longer training time, as RLW achieves on average its best validation $\Delta_m$ score at epoch $486.6$ (see Table~\ref{tab:city_results_main}). For the Scalarization approach, results for mIoU / PxAcc / AbsErr are reported in the following ranges: 0.0685-0.7025 / 0.915-0.92 / 0.0127-0.0133, which is also worse than our results for ResNet-50 shown in Table~\ref{tab:city_results_main}. Also, note that \cite{xin2022current} did not take into account the Relative Depth Error (RErr), which is a very sensitive metric and difficult to optimize.
For CelebA, they use the ResNet-18 as well with 100 epochs and report the lowest Average Error for EW at $9.23\%$ and for RLW (Normal) at $9.30\%$ (see~\cite[Figure 10]{xin2022current}. In our results in Table~\ref{tab:celebA_results_main}, we both improve upon these results. We argue that this is due to the choice of a different optimizer. While we use Adam, \cite{xin2022current} use SGD with momentum. Recent work \cite{eli2023_challenging} shows the superiority of Adam over SGD in MTL, which is most likely the reason for our improved results.

\subsection{Weaknesses of UW}
\label{sec:appx_weak_uw}
\begin{figure}
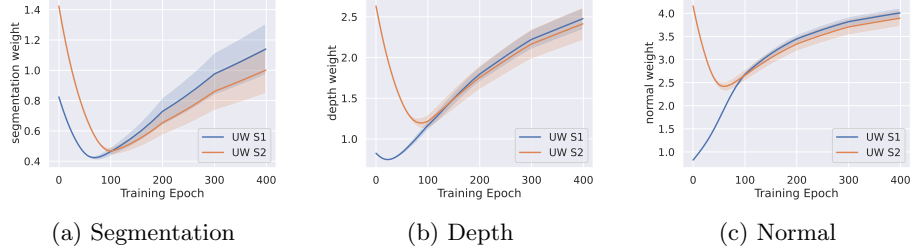

     \centering
     \begin{subfigure}[t]{0.3\textwidth}
         \centering
         \includegraphics[trim = 0mm 0mm 0mm 0mm, clip, width=\textwidth]{figures/UW_inertia_segmentation_talk.pdf}
         \caption{Segmentation}
     \end{subfigure}
     \hfill
     \begin{subfigure}[t]{0.3\textwidth}
         \centering
         \includegraphics[trim = 0mm 0mm 0mm 0mm, clip, width=\textwidth]{figures/UW_inertia_depth_talk.pdf}
         \caption{Depth}
     \end{subfigure}
     \hfill
    \begin{subfigure}[t]{0.3\textwidth}
         \centering
         \includegraphics[trim = 0mm 0mm 0mm 0mm, clip, width=\textwidth]{figures/UW_inertia_normal_talk.pdf}
         \caption{Normal}
     \end{subfigure}
     \hfill
    \caption{Comparison of the learning procedure of task weights for two different initializations of $\sigma$ for UW. Note that the plots do not show the raw $\sigma$ values, but the actual task weights $\frac{1}{2\sigma^2}$ that are ultimately determined by $\sigma$. Equal starting parameter values in blue (UW S1) following \cite{LibMTL}; learned parameters (from the final epoch from a previous UW S1 run) as starting values in orange (UW S2). We show the total task weights for a) semantic segmentation, b) depth estimation, and c) surface normal prediction on the NYUv2 dataset trained with the SegNet. We plot the mean task weight of 5 random seeds with the standard deviation as shaded area.
    }
    \label{fig:appx_weakness_UW}
\end{figure}

\begin{figure}
     \centering
     \begin{subfigure}[t]{0.4\textwidth}
         \centering
         \includegraphics[trim = 0mm 0mm 0mm 0mm, clip, width=\textwidth]{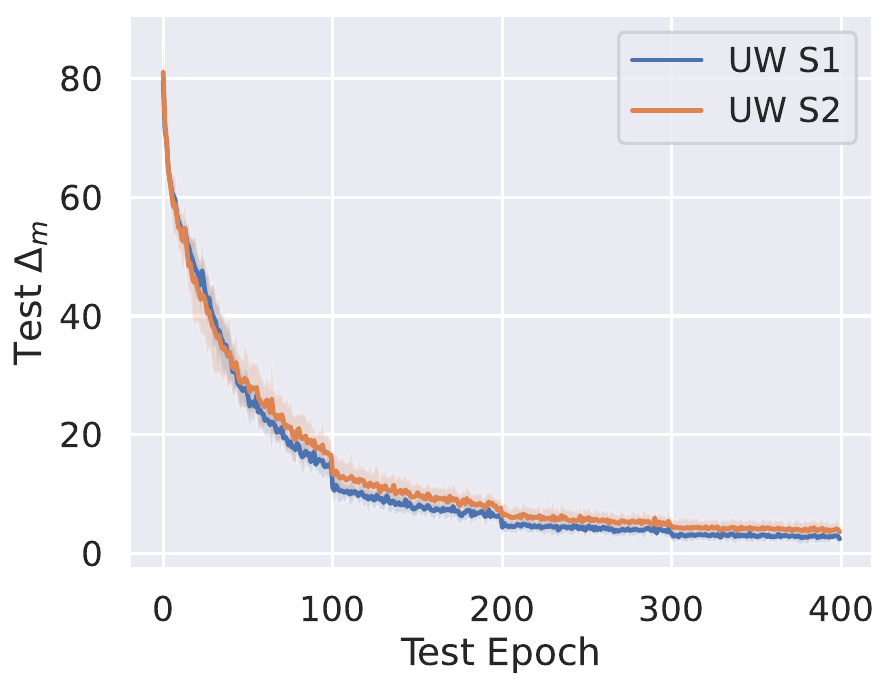}
         \caption{Test $\Delta_m$}
     \end{subfigure}
     \hfill
     \begin{subfigure}[t]{0.4\textwidth}
         \centering
         \includegraphics[trim = 0mm 0mm 0mm 0mm, clip, width=\textwidth]{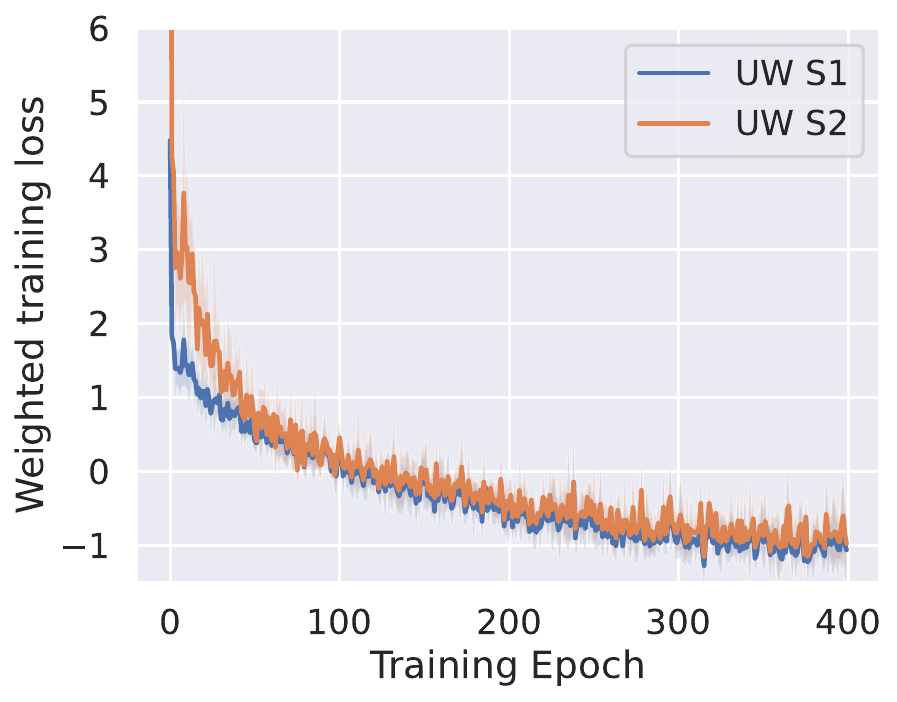}
         \caption{Training loss}
     \end{subfigure}
     \hfill
    \caption{For the same setup as in Figure~\ref{fig:appx_weakness_UW} the (a) Test $\Delta_m$ is displayed and (b) the overall weighted training loss. The Test $\Delta_m$ for UW S2 is worse than for UW S1, probably due to the \textit{inertia} phenomena. The weighted training loss is obviously higher for the first approximately 90 epochs, until weights for UW S2 (Figure~\ref{fig:appx_weakness_UW}) are on  a similar level. But overall, the loss for UW S2 stays a bit higher than for UW S1 - maybe because the learning was not so effective in the first 90 epochs compared to UW S2 - which might be the reason for the worse Test $\Delta_m$ score. }
    \label{fig:appx_weakness_UW_loss}
\end{figure}

\paragraph{Inertia}
In Figure \ref{fig:appx_weakness_UW}, we show how UW suffers from high inertia combined with a sub-optimal initialization. Only by initializing $\sigma$ differently, the gradient updates need almost one-fourth of the total training time to compensate for this. This results in a worse overall performance as can be observed in Figure~\ref{fig:appx_weakness_UW_loss}. In particular, the gap between the training loss of the two different experiments UW S1 and UW S2 indicates that the initialization has an impact on the model performance. However, the problem is that it is difficult to get proper initial weights for UW. Thus, usually an equal task weight initialization is applied, which might not be optimal. Taking the analytically optimal solution in each iteration instead of using gradient descent circumvents the inertia (see our solution in Section \ref{sec_uwso}).

\paragraph{Model complexity}
Adding to the observations of the authors of UW \cite{kendall2018multi} who claim that it only depends on aleatoric homoscedastic uncertainty, we empirically prove that UW additionally depends on the model complexity. The dependence on the homoscedastic uncertainty means that the authors hypothesize that when training a neural network with UW, $\sigma$ converges towards the standard deviation of the aleatoric uncertainty (i.e., inherent noise present in the data such as sensor noise), assuming we have infinite data and infinite model capacity. Using a simple toy example, we can prove that this holds true. For this, we train a simple neural network with three hidden layers and a linear output layer to learn the function $f(x) = x + \epsilon$, where $\epsilon$ is the data noise. The noise is sampled in each training iteration from a normal distribution with $\mu=0$ and added to the continuous inputs $x$ which have values between 0 and 1. $\sigma$ converges towards the standard deviation of the noise when increasing the amount of data. We can observe the same learning behavior when we replace our linear toy function with a slightly more complex one that also includes kinks.

However, when we increase the complexity of the function or data and use the same model, the hypothesis of \cite{kendall2018multi} does not hold anymore. When training a complex function using a small model complexity, we can observe that UW $\sigma$ converges to a value that is greater than the standard deviation of the noise. The difference between std dev and $\sigma$ reduces with increasing training data, i.e., we reduce uncertainty by adding more data. Table \ref{table:uw_model_complexity} shows the values of $\sigma$ in the last training iteration for different standard deviations and an increasing number of data points. While $\sigma$ is approximately equal to the noise std dev with a simple function trained with only 100 data points, this equality does not hold for a more complex function, even with $10^7$ data points.
Therefore, we conclude that UW does not only depend on aleatoric homoscedastic uncertainty (Kendall et al. \cite{kendall2018multi}), but also \textbf{depends on model complexity.}

\begin{table}[ht!]
\centering
\caption{Learned UW $\sigma$ values (last training iteration) for different noise standard deviations / amounts of data using a complex function and a simple neural network. $\sigma$ does not converge towards the noise std dev even with a large amount of training data. Instead, UW $\sigma$ is greater than the std dev of the data noise.}
\begin{tabular}{l|l|lllll}
\toprule
 &  & \multicolumn{4}{c}{Number of data points} \\
 &  & $10^2$ & $10^3$ & $10^4$ & $10^6$ & $10^7$ \\ 
 \midrule
\multirow{4}{*}{Noise std} & 0.0 & 0.817 & 0.790 & 0.291 & 0.203 & 0.173 \\
 & 0.1 & 0.814 & 0.784 & 0.321 & 0.195 & 0.196 \\
 & 0.2 & 0.816 & 0.816 & 0.934 & 0.290 & 0.222 \\
 & 0.3 & 0.817 & 0.820 & 0.529 & 0.388 & 0.331 \\
 \bottomrule
\end{tabular}

\label{table:uw_model_complexity}
\end{table}

\begin{figure}
    \centering
    \includegraphics[clip, width=1.0\columnwidth]{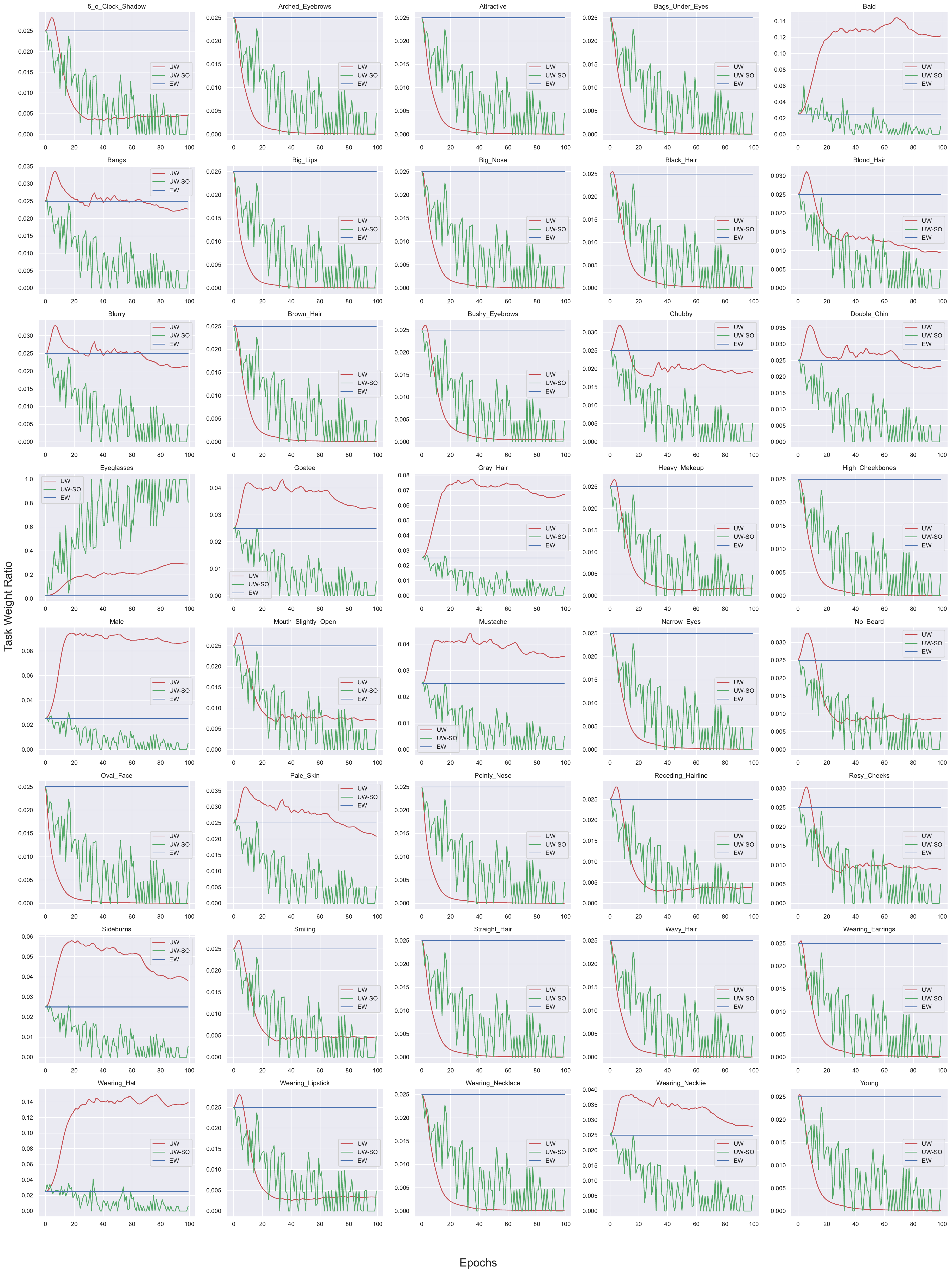}
    \caption{Comparison of the weight ratios of UW, UW-SO, and EW on the CelebA dataset using the ResNet-18. }
    \label{fig:appx_celebA_weights}
\end{figure}

\begin{figure}
    \centering
    \includegraphics[clip, width=1.0\columnwidth]{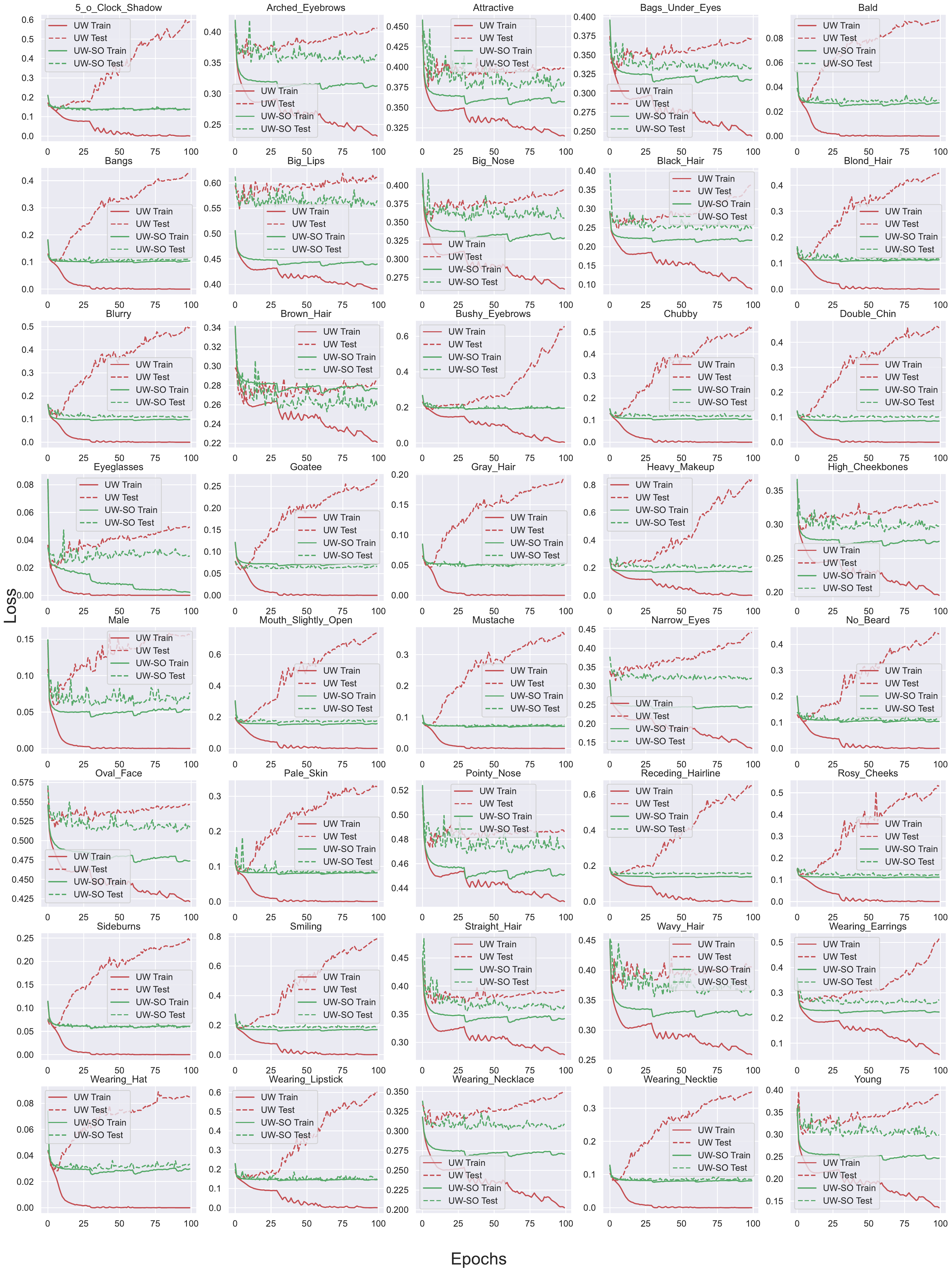}
    \caption{Comparison of the losses of UW and UW-SO on the CelebA dataset using the ResNet-18. }
    \label{fig:appx_celebA_losses}
\end{figure}

\paragraph{Overfitting of CelebA}
For CelebA, we encountered severe overfitting problems of UW. To better understand the cause, we look at the weight ratio (see Figure~\ref{fig:appx_celebA_weights}) as well as the loss (see Figure~\ref{fig:appx_celebA_losses}) for each of the 40 tasks in CelebA, and compare it to our loss weighting method UW-SO which mitigates the overfitting problem.

Many tasks show the behavior described for the \textit{Bald} task in Section \ref{sec:overfitting} where there is a huge gap between train and test loss caused by a high task weight. For instance, \textit{Male / Wearing Hat / Gray Hair} have relative weights of about $0.09, 0.14, 0.07$ already within the first 20 epochs (in comparison EW constantly assigns $0.025$). We observe that UW-SO also has a strong focus on the single task \textit{Eyeglasses}, however, this happens later in training and task weights are fluctuating stronger, thus, each task is assigned a proper weight occasionally.

\paragraph{Inertia of UW for CelebA}
When looking at the weight ratios for the 40 tasks of CelebA (see Figure~\ref{fig:appx_celebA_weights}), we can also encounter some initialization and inertia effects. For example, tasks such as \textit{Sideburns} in row 7, column 1 (r7, c1), \textit{Wearing Hat} (r8, c1), \textit{Gray Hear} (r4, c3), \textit{Bald} (r1, c5), or \textit{Male} (r5, c1) have a steady and slow weight increase within the first 20 epochs. Thus, it takes almost one-fifth of the training time to adapt the weight to a level that is desired by UW.

\clearpage

\subsection{Weaknesses of Scalarization}
\label{sec:appx_scalar_search}
To demonstrate the high computational cost of the brute force Scalarization approach, we show the resulting experiments required to perform the grid search of learning rate and scalar weights for the NYUv2 dataset in Figure~\ref{fig:appx_lr_wd}. Note that this is part of our line search hyperparameter optimization approach, where we perform a grid search over learning rate and scalars using a fixed weight decay. Thus, the results are slightly worse than those reported in Table~\ref{tab:nyuv2_results_main}, as the weight decay is not yet optimized. Also, note that we only show the results for 3 different learning rates, including the best one. When looking at 3 learning rates in combination with 3 different scalars $\omega_i$ (one for each task), where $\sum_i \omega_i = 1$, we have to perform 108 different runs (36 weight combinations per learning rate), which is computationally highly expensive compared to weighting methods that do not include hyperparameters, such as UW. Also, only 3 out of 108 experiments result in a $\Delta_m$ score below 0. Thus, finding the optimum using Scalarization requires careful tuning of task weights, which becomes computationally intractable and practically not feasible for a large number of tasks.

Furthermore, the choice of step size is non-trivial and depends on the dataset. While a step size of $0.1$ achieves already good results with NYUv2, it is no fine enough for Cityscapes. Thus, we had to tune the weights in Cityscapes with a step size of $0.02$ to get proper results.

\subsection{Toy Example: Scaled NYUv2}
\label{appx:motivation_uwo}
\begin{figure}[ht]
     \centering
     \begin{subfigure}[t]{0.3\textwidth}
         \centering
         \includegraphics[trim = 0mm 0mm 0mm 0mm, clip, width=\textwidth]{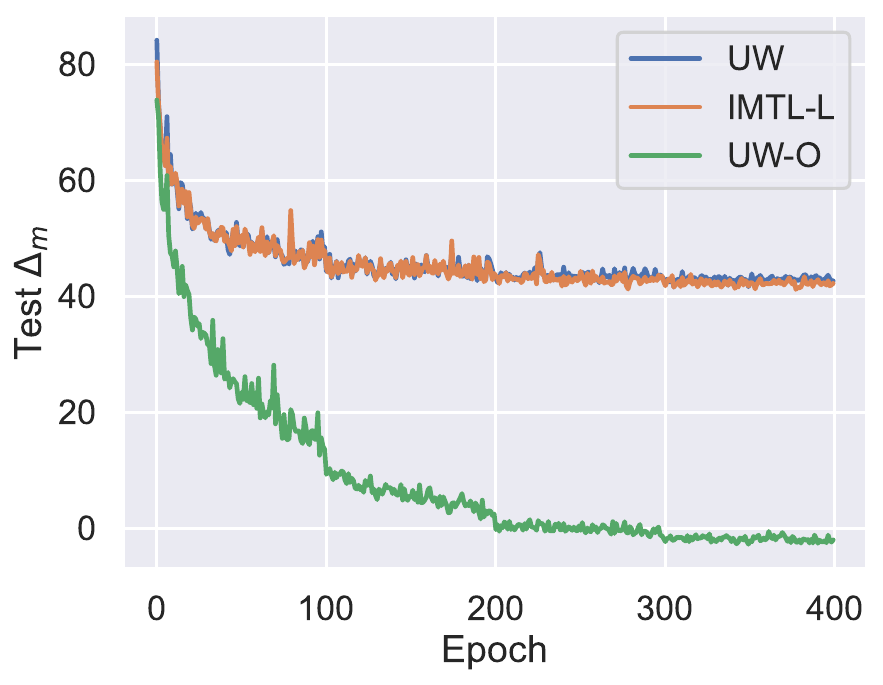}
         \caption{Test $\Delta_m$}
     \end{subfigure}
     \hfill
     \begin{subfigure}[t]{0.3\textwidth}
         \centering
         \includegraphics[trim = 0mm 0mm 0mm 0mm, clip, width=\textwidth]{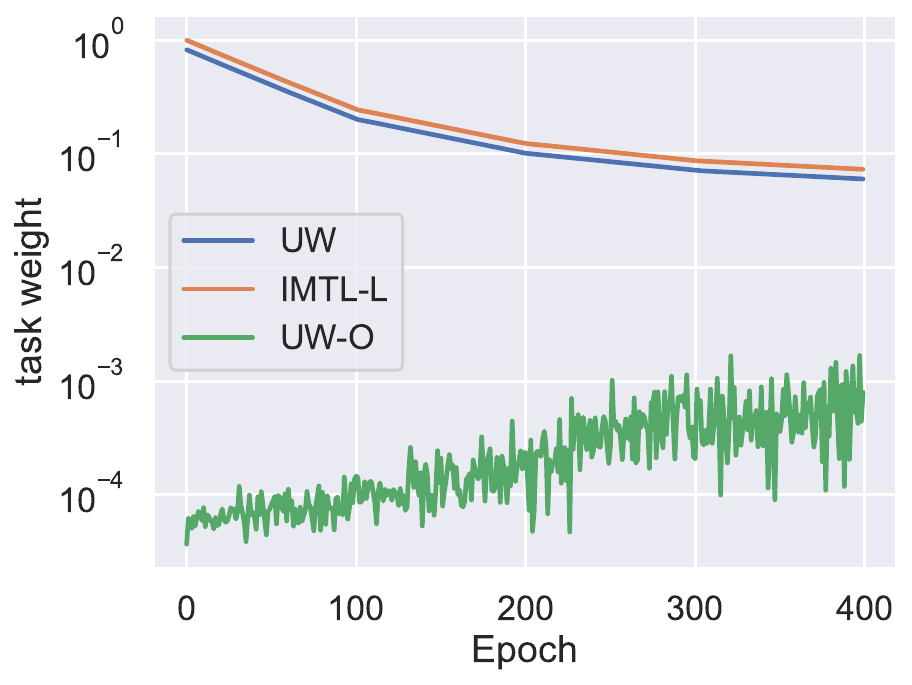}
         \caption{Segmentation task weight}
     \end{subfigure}
     \hfill
     \begin{subfigure}[t]{0.3\textwidth}
         \centering
         \includegraphics[trim = 0mm 0mm 0mm 0mm, clip, width=\textwidth]{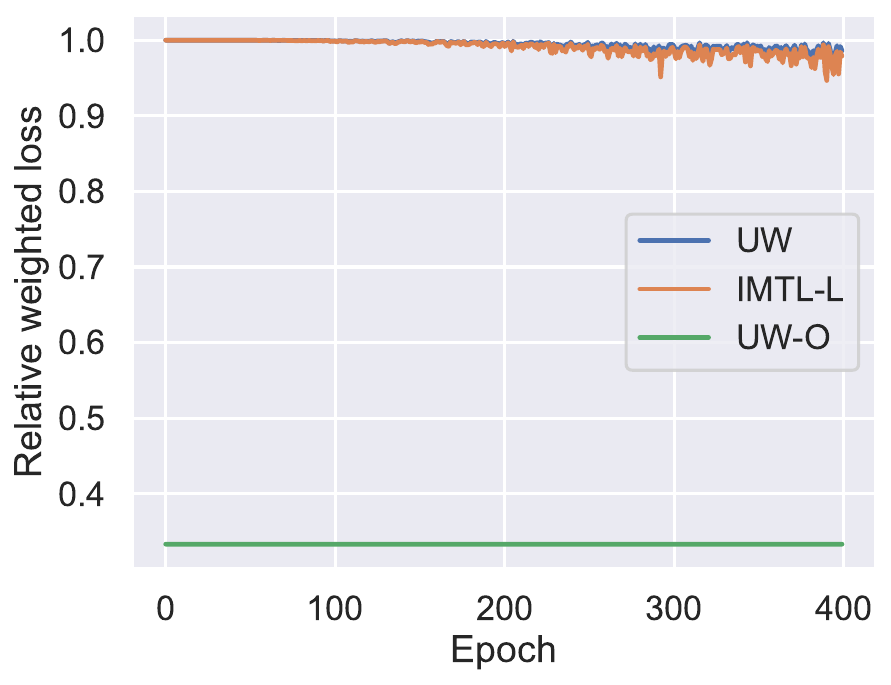}
         \caption{Relative weighted segmentation loss}
     \end{subfigure}
     \hfill
    \caption{Comparison of UW, IMTL-L, and UW-O on the scaled NYUv2 toy example trained with the SegNet in terms of a) test $\Delta_m$, b) segmentation task weight development, and c) relative weighted segmentation loss. We show the segmentation task in particular as it is scaled with $10,000$. Thus, finding a proper balance for this task is the difficulty in this scaled toy example. While UW and IMTL-L fail to minimize the $\Delta_m$ score, UW-O seems to find a good solution for all tasks as indicated by the much smaller $\Delta_m$ score. This is because UW-O assigns a very small weight (between $10^{-4}$ and $10^{-3}$ to the highly scaled segmentation task, while UW and IMTL-L suffer from inertia and cannot reduce the segmentation task weight accordingly. Therefore, the relative weighted loss for the segmentation task using UW or IMTL-L results in nearly $100\%$, while UW-O normalizes each loss by definition.
    }
    \label{fig:appx_scaled_nyu}
\end{figure}

We perform a toy experiment in which we scale the three losses (segmentation, depth, normal) of the NYUv2 dataset with $10,000$, $0.1$, and $1.0$, respectively, and compare the $\Delta_m$ test data performance of four weighting methods across different architectures in Table~\ref{tab:nyuv2_scaled_summary}. 
Only GLS and UW-O are unaffected by the scaling and achieve a similar result as for the unscaled dataset. Small differences to the unscaled results (compare Table~\ref{tab:nyuv2_results_main} for full results) can also be attributed to only one random seed having been used. UW-O is independent of loss magnitudes by definition, whereas the similar method IMTL-L which tries to scale 
each weighted task loss $\omega_k L_k$ to $1$ by learning $\omega_k$ (see Section~\ref{sec_uwso}) is considerably worse as it is (as UW) too slow in adapting the weight for the large segmentation loss (see Figure~\ref{fig:appx_scaled_nyu}). 
Thus, UW-O offers an easy and well-performing alternative to handle highly imbalanced losses. In contrast to GLS, it can be applied independently of the number of tasks. GLS could, e.g., not be applied for the 40-task CelebA experiments as 40 losses with each loss $L_k < 0$ results in a multiplied loss under the square root too small to be computationally tractable.
Interestingly, as Table~\ref{tab:nyuv2_scaled_summary} shows, the negative impact of the massively scaled losses on IMTL-L and UW decreases with larger networks.

\begin{table}[t!]
\centering
\small
\scriptsize 
\setlength\tabcolsep{2pt} 
\caption{Values for test $\Delta_m$ for the NYUv2 toy experiment with large loss magnitude differences (scaled) for different architectures over four MTO methods on one random seed. The best score is highlighted in bold. We compare results with the unscaled experiments from Table~\ref{tab:nyuv2_results_main}. Only GLS and UW-O are not affected by large-scale task loss differences.}
\begin{tabular}{lrrrr}
\toprule
          & GLS   & IMTL-L & UW   & UW-O  \\
\midrule
SegNet unscaled    & 2.4   & -0.3   & -0.3 & -0.0  \\
\rowcolor{Gray}
SegNet scaled    & 0.6   & 41.2   & 41.9 & \pmb{-2.5}  \\
ResNet-50 unscaled  & -8.7  &   -7.7  & -7.2  & -9.1  \\
\rowcolor{Gray}
ResNet-50 scaled  & \pmb{-8.7}  & 0.4    & 2.4  & -8.5  \\
ResNet-101 unscaled & -11.6 &  -10.3   & -10.4  & -11.9 \\
\rowcolor{Gray}
ResNet-101 scaled & -11.5 & 2.1    & 0.3  & \pmb{-12.6} \\
\bottomrule
\end{tabular}

\label{tab:nyuv2_scaled_summary}
\end{table}


\subsection{Comparison of Task Weights}
\label{sec:appx_task_weights}
\begin{figure}[ht]
     \centering
     \begin{subfigure}[t]{0.3\textwidth}
         \centering
         \includegraphics[trim = 0mm 0mm 0mm 0mm, clip, width=\textwidth]{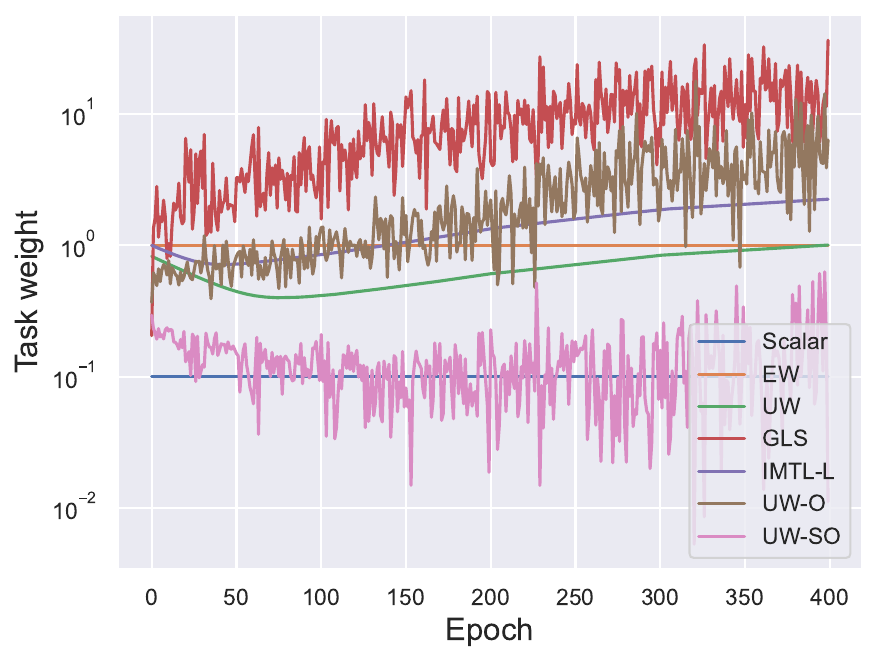}
         \caption{Segmentation}
     \end{subfigure}
     \hfill
     \begin{subfigure}[t]{0.3\textwidth}
         \centering
         \includegraphics[trim = 0mm 0mm 0mm 0mm, clip, width=\textwidth]{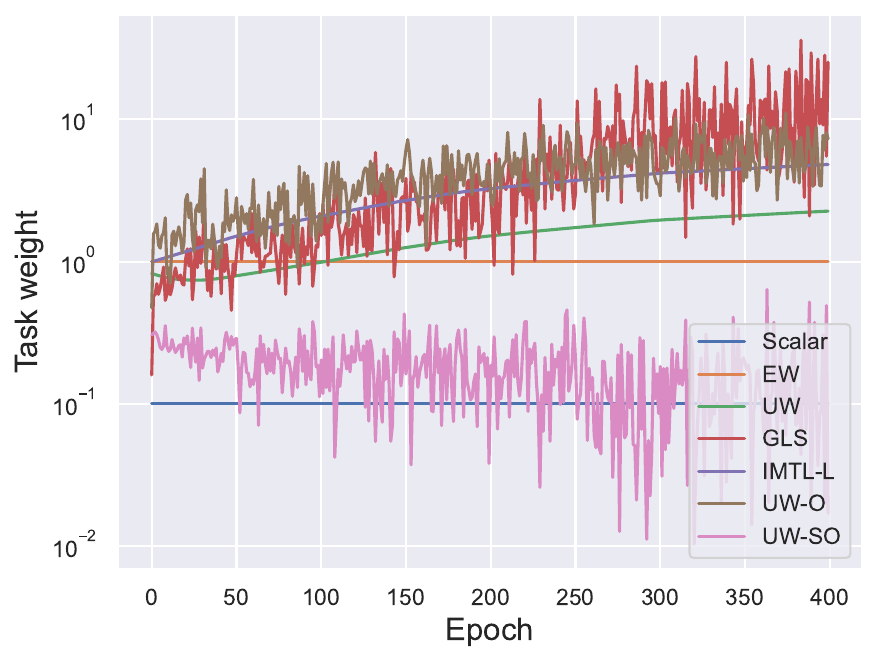}
         \caption{Depth}
     \end{subfigure}
     \hfill
    \begin{subfigure}[t]{0.3\textwidth}
         \centering
         \includegraphics[trim = 0mm 0mm 0mm 0mm, clip, width=\textwidth]{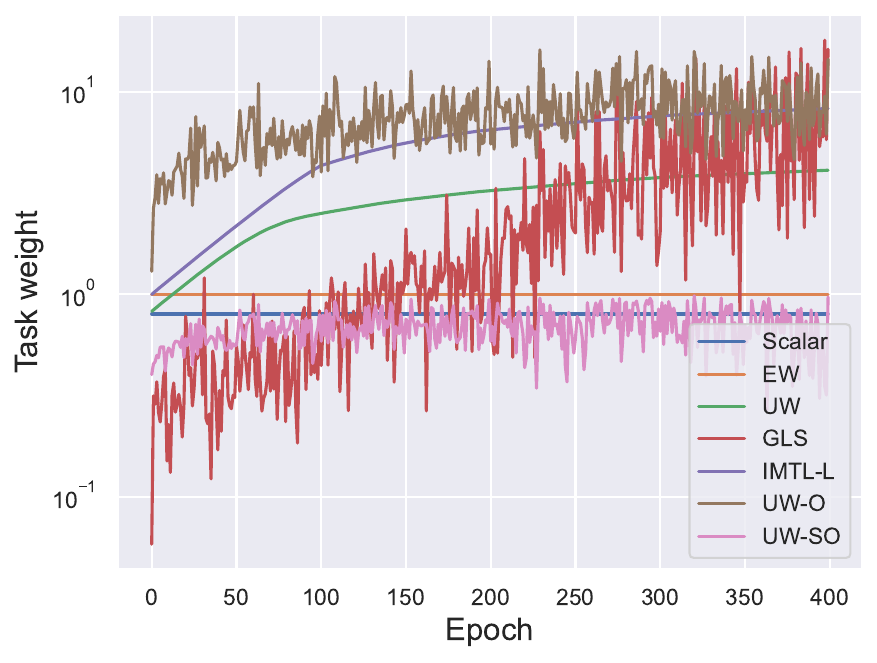}
         \caption{Normal}
     \end{subfigure}
     \hfill
    \caption{Weight development of various MTO algorithms for the NYUv2 tasks using the SegNet.
    }
    \label{fig:appx_nyu_weights}
\end{figure}

\begin{figure}[ht]
    \centering
    \begin{subfigure}{.3\textwidth}
        \centering
        \includegraphics[width=\textwidth,keepaspectratio]{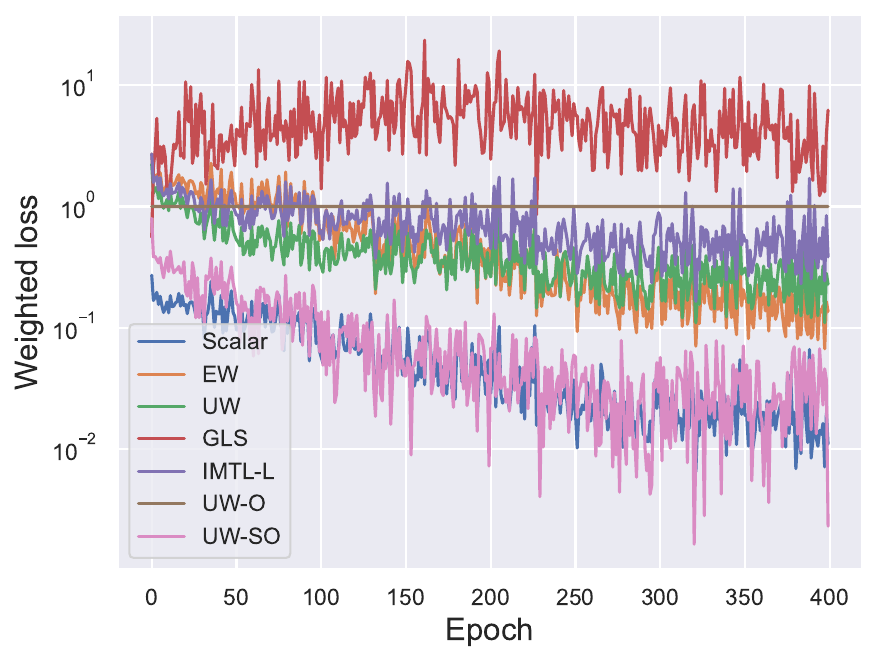}  
        \caption{Segmentation}
        \label{fig:nyu_lr_weighted}
    \end{subfigure}
    \hfill
    \begin{subfigure}{.3\textwidth}
        \centering
        \includegraphics[width=\textwidth,keepaspectratio]{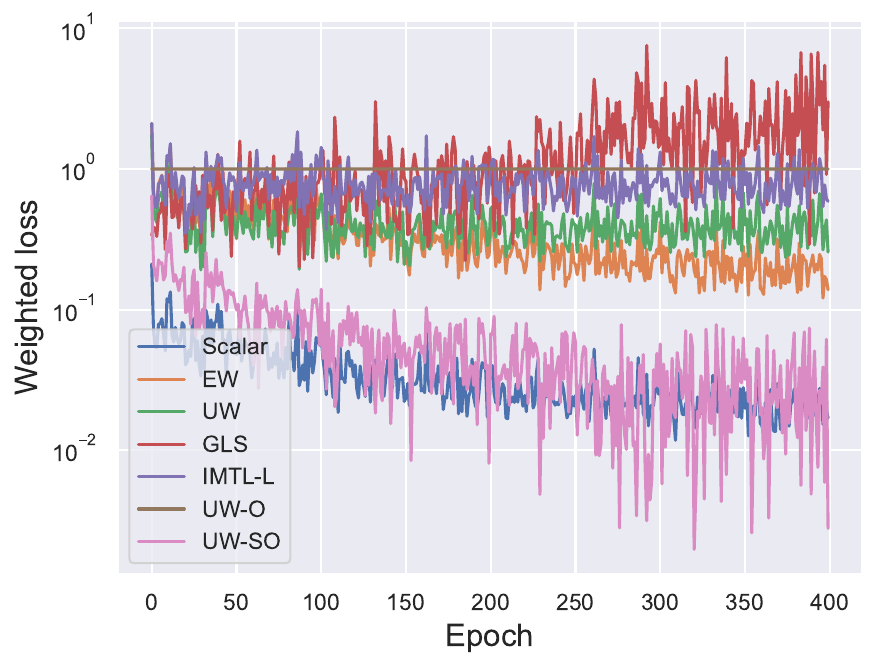}
        \caption{Depth}
        \label{fig:city_lr_weighted}
    \end{subfigure}
    \hfill
    \begin{subfigure}{.3\textwidth}
        \centering
        \includegraphics[width=\textwidth,keepaspectratio]{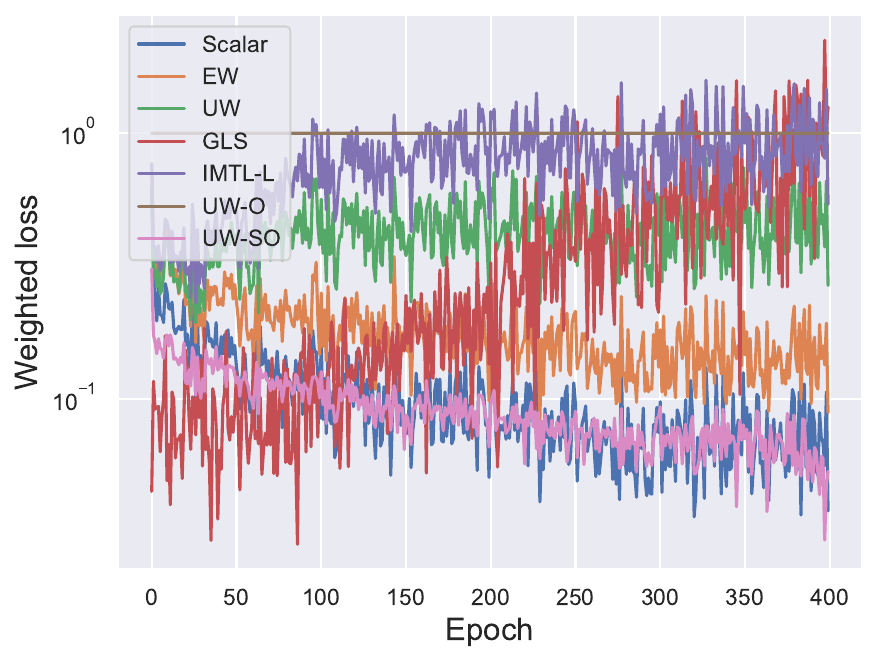}  
        \caption{Normal}
        \label{fig:celebA_lr_weighted}
    \end{subfigure}
    \hfill
    \caption{Weighted loss development of various MTO algorithms for the NYUv2 tasks using the SegNet.}
  \label{fig:appx_nyu_weighted_loss}
\end{figure}

\begin{figure}[ht]
     \centering
     \begin{subfigure}[t]{0.3\textwidth}
         \centering
         \includegraphics[trim = 0mm 0mm 0mm 0mm, clip, width=\textwidth]{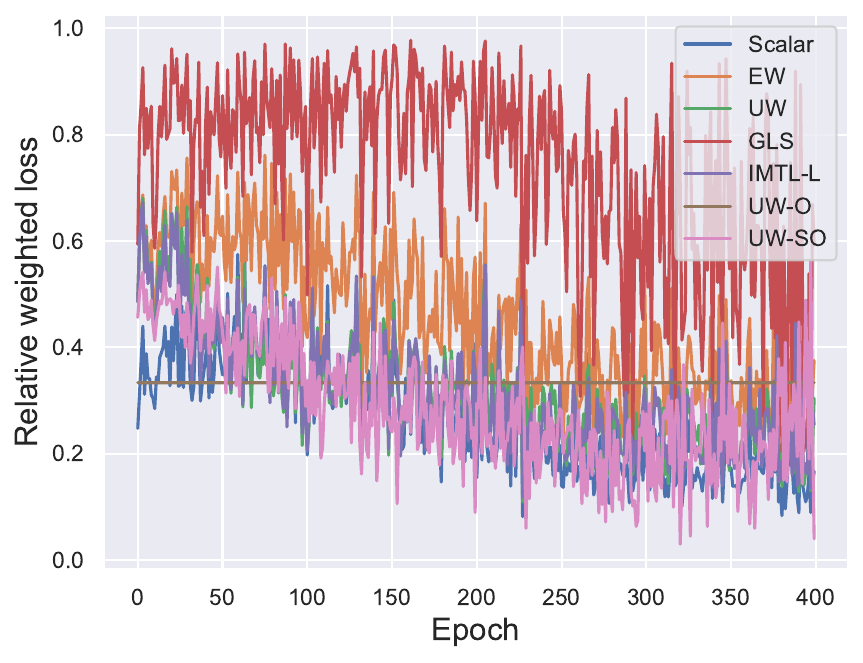}
         \caption{Segmentation}
     \end{subfigure}
     \hfill
     \begin{subfigure}[t]{0.3\textwidth}
         \centering
         \includegraphics[trim = 0mm 0mm 0mm 0mm, clip, width=\textwidth]{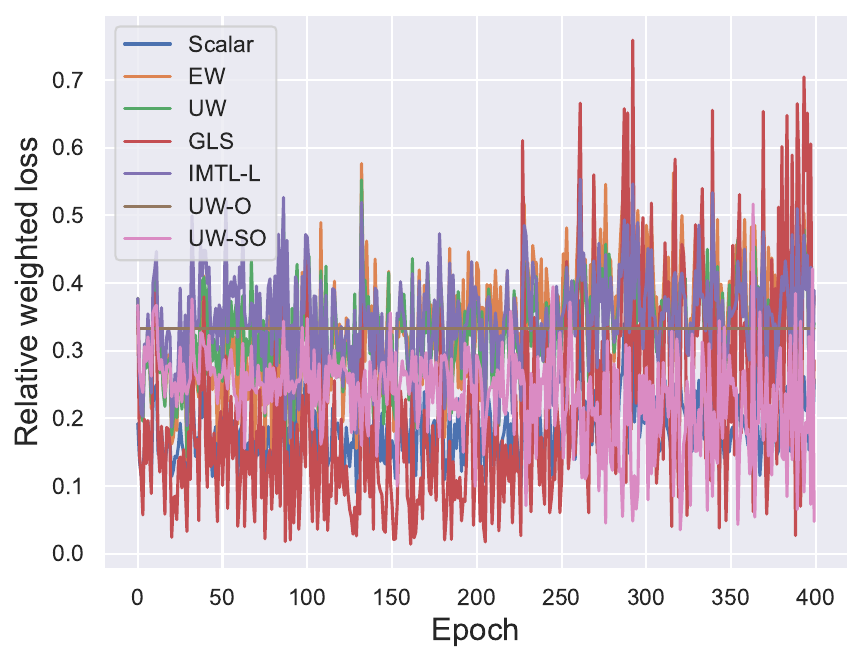}
         \caption{Depth}
     \end{subfigure}
     \hfill
    \begin{subfigure}[t]{0.3\textwidth}
         \centering
         \includegraphics[trim = 0mm 0mm 0mm 0mm, clip, width=\textwidth]{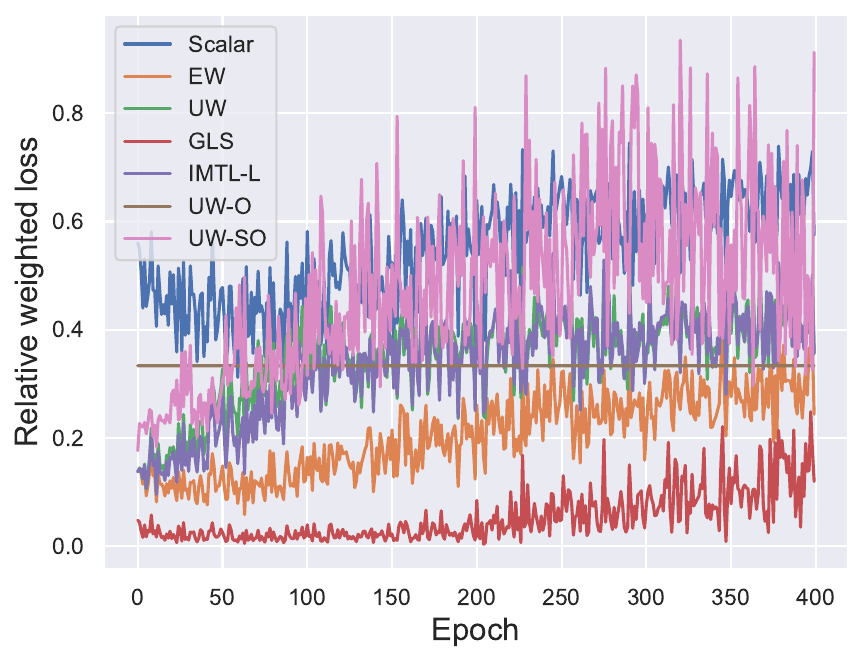}
         \caption{Normal}
     \end{subfigure}
     \hfill
    \caption{Relative weighted loss development of various MTO algorithms for the NYUv2 tasks using the SegNet.}
    \label{fig:appx_nyu_rel_weights}
\end{figure}

Comparing the task weights along the course of training for various weighting methods allows us to better understand the respective weighting principles of each MTO approach. Figure~\ref{fig:appx_nyu_weights} exemplarily shows the weights for the 3 tasks of the NYUv2 dataset. Furthermore, Figures~\ref{fig:appx_nyu_weighted_loss} and \ref{fig:appx_nyu_rel_weights} represent the weighted losses and the relative weighted losses, respectively. Besides the similarity of Scalarization and UW-SO as already described in Section~\ref{sec:comparison_common_lw_methods}, we can also observe that UW-O and IMTL-L show a fairly similar task weight development. While IMTL-L develops the weights rather smoothly due to the small gradient updates, UW-O shows more fluctuation and jitters around IMTL-L. This again shows that both methods have the same underlying objective (scale losses to 1), only the derivation is performed differently as IMTL-L learns the weights using gradient descent and UW-O computes the weights based on the task losses.

\subsection{Oscillation of MTO methods}

The \textbf{weight development} (Figure~\ref{fig:appx_nyu_weights}) of UW and IMTL-L is similar, as the task weights first decrease before they increase steadily. Both methods are gradient-based w.r.t. the update of the task weights and show a smooth change of weights. Looking at loss-based MTO methods, like GLS, UW-O, UW-SO, we see that task weights fluctuate more. Constant weighting methods, like EW and Scalar show no oscillation, by definition.

Looking at the \textbf{weighted loss} (Figure~\ref{fig:appx_nyu_weighted_loss}) we see no difference among the methods in the magnitude of oscillation with the exception of the UW-O method which by construction has a constant weighted loss.

\subsection{Influence of learning rate and weight decay}
In our experiments, we find that the LR has a high impact on the performance while the influence of the WD is low in comparison (see Figures~\ref{fig:abl_lr} and \ref{fig:appx_lr_wd}). This underlines our line search approach to first tune the LR (with any method hyperparameters, if present) and then tune the WD. We show that this does not worsen results compared to a full grid search (see Section \ref{sec:appx_impl_details}). Thus, for practical applications, tuning the LR is crucial, but tuning the WD seems to be less important, especially for more complex networks.

\subsection{Search for temperature $T$}

\begin{figure}[ht]
     \centering
     \begin{subfigure}[t]{0.3\textwidth}
         \centering
         \includegraphics[trim = 0mm 0mm 0mm 0mm, clip, width=\textwidth]{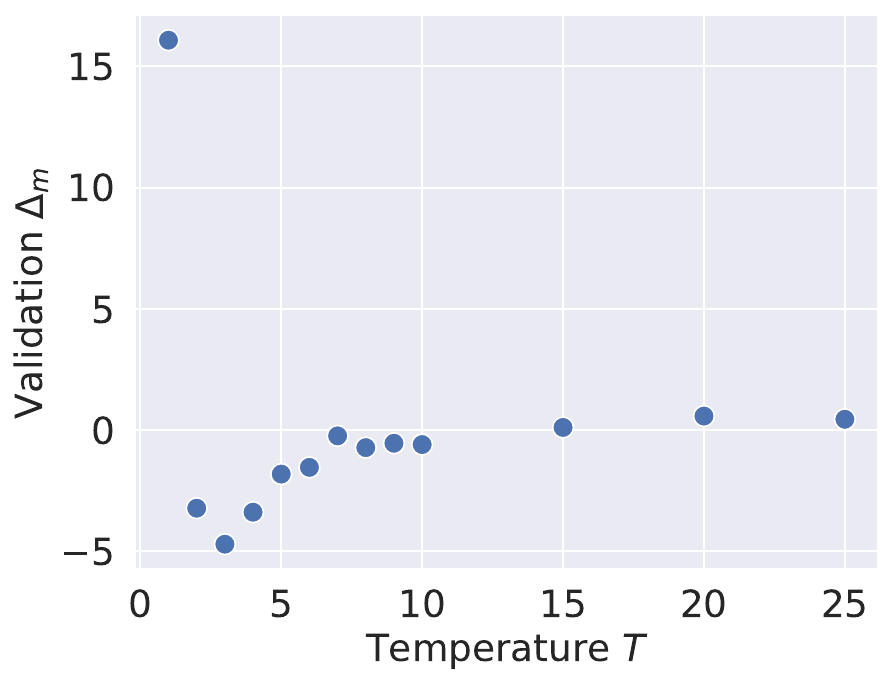}
         \caption{SegNet}
     \end{subfigure}
     \hfill
     \begin{subfigure}[t]{0.3\textwidth}
         \centering
         \includegraphics[trim = 0mm 0mm 0mm 0mm, clip, width=\textwidth]{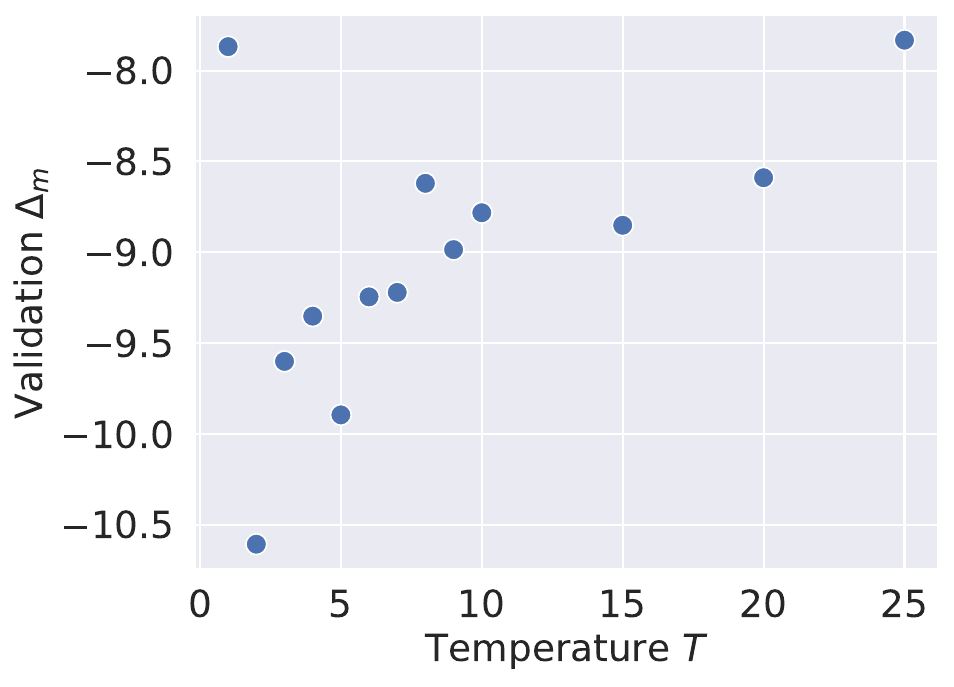}
         \caption{ResNet-50}
     \end{subfigure}
     \hfill
    \begin{subfigure}[t]{0.3\textwidth}
         \centering
         \includegraphics[trim = 0mm 0mm 0mm 0mm, clip, width=\textwidth]{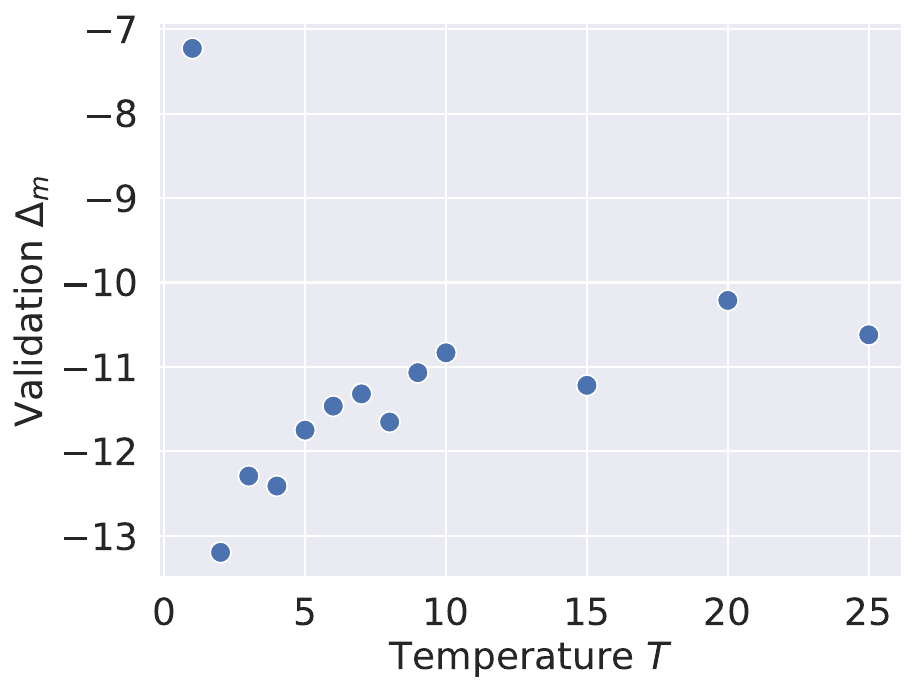}
         \caption{ResNet-101}
     \end{subfigure}
     \hfill
    \caption{Final $\Delta_m$ on NYUv2 performance of UW-SO for different choices for temperature $T$ using different networks.
    We use an initial step size of 5
    and refine the step size in the found area of interest between 1 and 10.
    }
    \label{fig:appx_abl_T_nyu}
\end{figure}

We show the final $\Delta_m$ on NYUv2 for different temperatures $T$ using our 2-stage search grid in Figure~\ref{fig:appx_abl_T_nyu}.

\begin{figure}[ht]
     \centering
     \begin{subfigure}[t]{0.32\textwidth}
         \centering
         \includegraphics[trim = 0mm 0mm 0mm 0mm, clip, width=\textwidth]{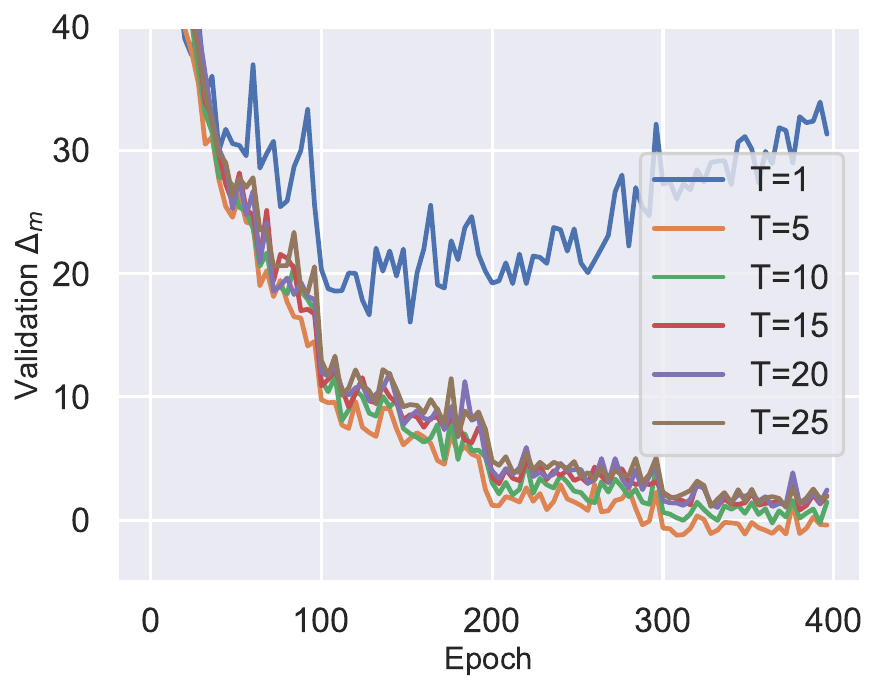}
         \caption{SegNet}
     \end{subfigure}
     \hfill
     \begin{subfigure}[t]{0.32\textwidth}
         \centering
         \includegraphics[trim = 0mm 0mm 0mm 0mm, clip, width=\textwidth]{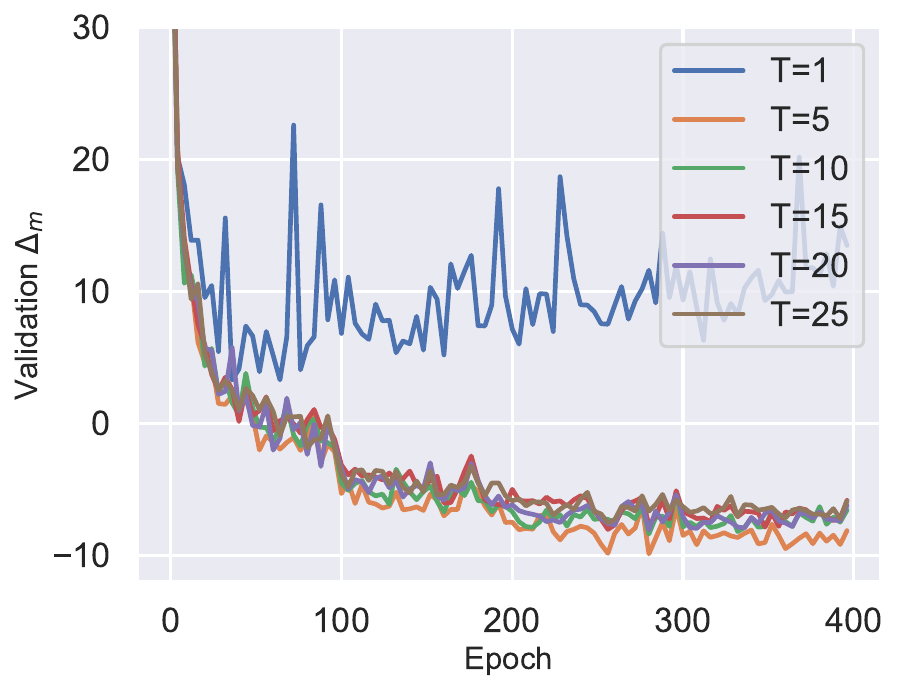}
         \caption{ResNet-50}
     \end{subfigure}
     \hfill
    \begin{subfigure}[t]{0.32\textwidth}
         \centering
         \includegraphics[trim = 0mm 0mm 0mm 0mm, clip, width=\textwidth]{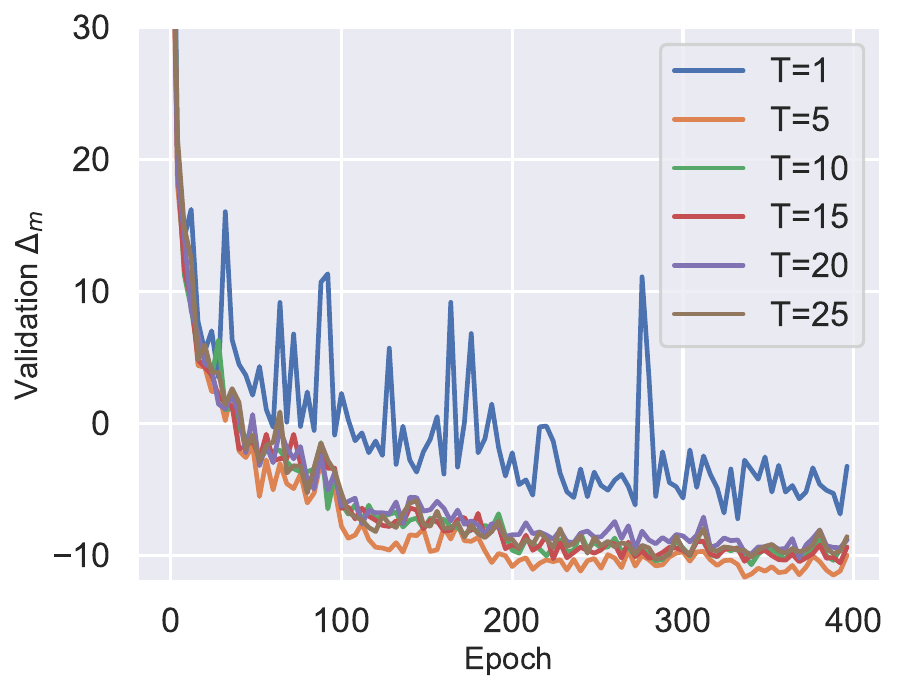}
         \caption{ResNet-101}
     \end{subfigure}
     \hfill
    \caption{Development of validation $\Delta_m$ for $T$ values with initial step size 5 for NYUv2 using the SegNet, ResNet-50, and ResNet-101.
    }
    \label{fig:appx_searchT_rough}
\end{figure}

\begin{figure}[ht]
     \centering
     \begin{subfigure}[t]{0.32\textwidth}
         \centering
         \includegraphics[trim = 0mm 0mm 0mm 0mm, clip, width=\textwidth]{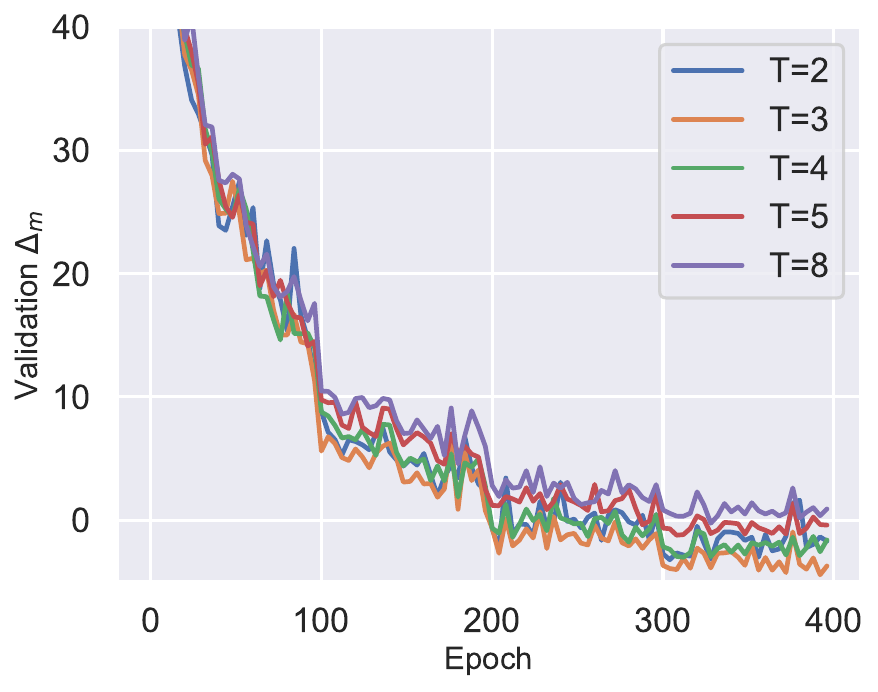}
         \caption{SegNet}
     \end{subfigure}
     \hfill
     \begin{subfigure}[t]{0.32\textwidth}
         \centering
         \includegraphics[trim = 0mm 0mm 0mm 0mm, clip, width=\textwidth]{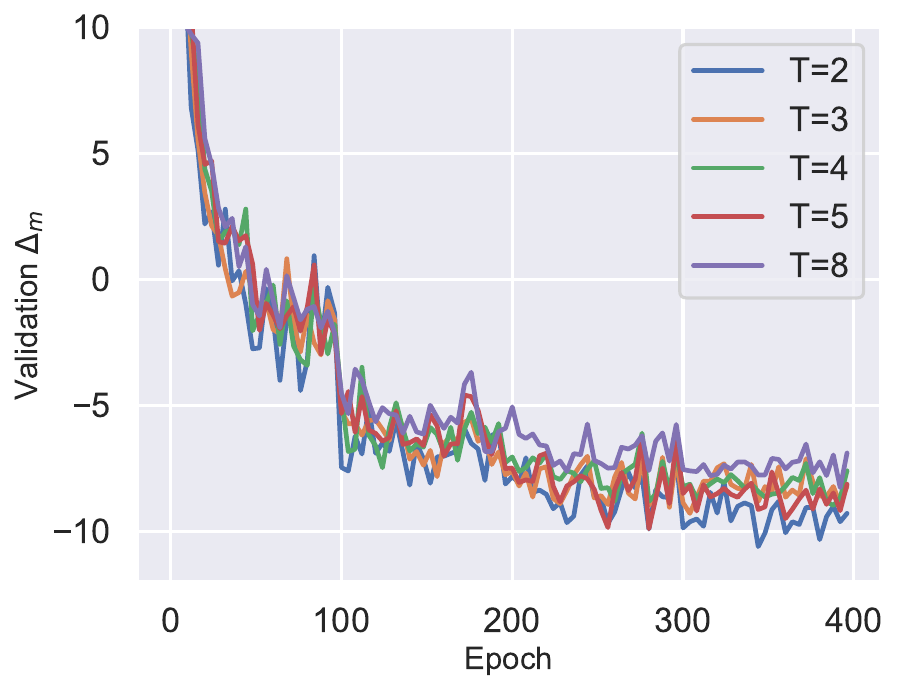}
         \caption{ResNet-50}
     \end{subfigure}
     \hfill
    \begin{subfigure}[t]{0.32\textwidth}
         \centering
         \includegraphics[trim = 0mm 0mm 0mm 0mm, clip, width=\textwidth]{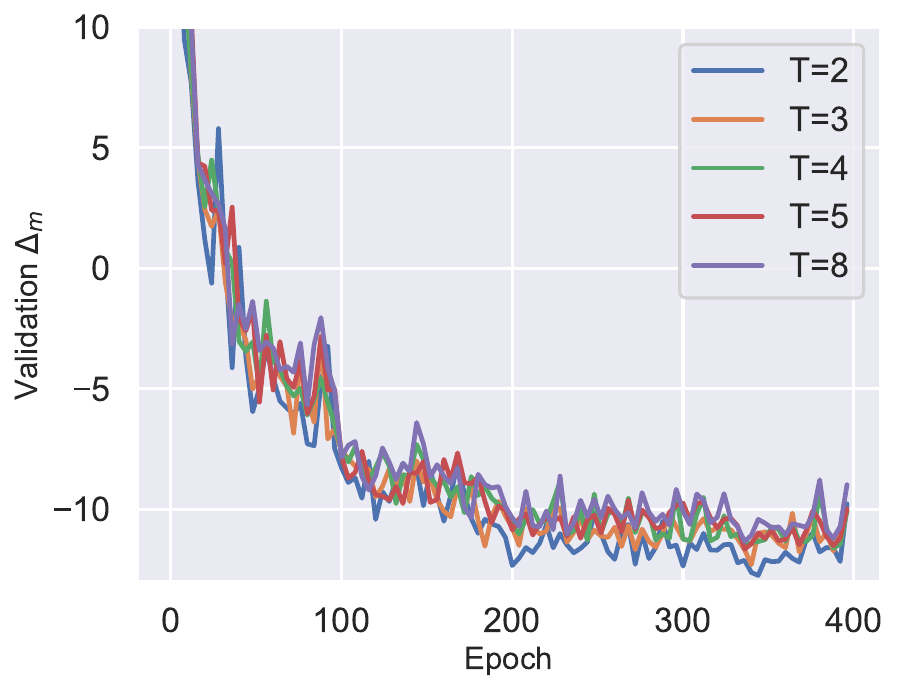}
         \caption{ResNet-101}
     \end{subfigure}
     \hfill
    \caption{Development of validation $\Delta_m$ for $T$ values with a more fine-grained step size for NYUv2 using the SegNet, ResNet-50, and ResNet-101.
    }
    \label{fig:appx_searchT_fine}
\end{figure}

\paragraph{Early Stopping of Non-optimal Configurations}
In Figures~\ref{fig:appx_searchT_rough} and \ref{fig:appx_searchT_fine} we show how the search for the softmax temperature $T$ for UW-SO can be simplified, as the optimal value becomes already visible after around one-fourth of the training time. Therefore, we suggest to follow a sequential search with an initial step size of 5 followed by a finer search around the previous optimum.

\begin{figure}[ht]
     \centering
     \begin{subfigure}[t]{0.32\textwidth}
         \centering
         \includegraphics[trim = 0mm 0mm 0mm 0mm, clip, width=\textwidth]{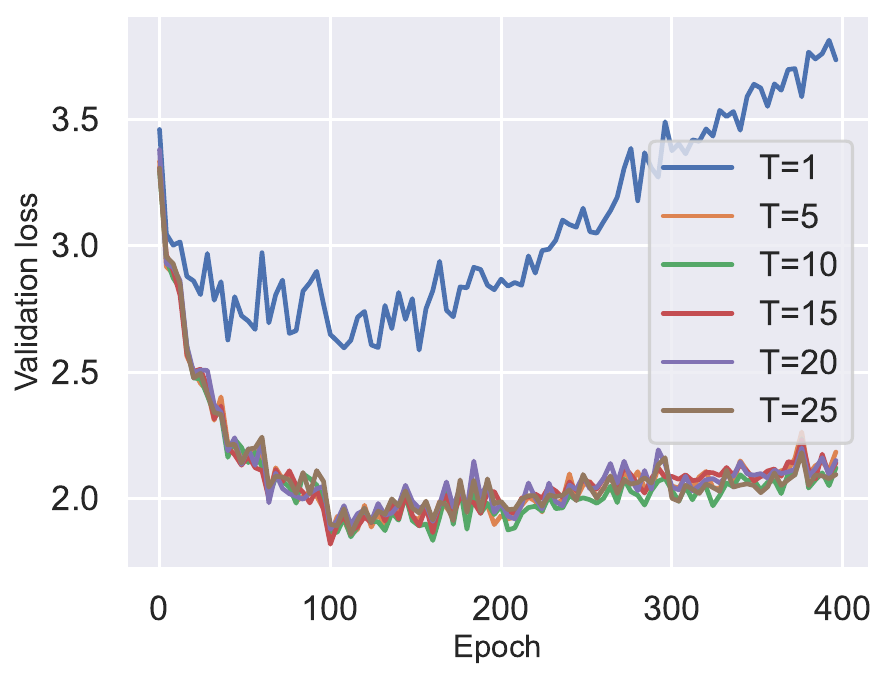}
         \caption{SegNet}
     \end{subfigure}
     \hfill
     \begin{subfigure}[t]{0.32\textwidth}
         \centering
         \includegraphics[trim = 0mm 0mm 0mm 0mm, clip, width=\textwidth]{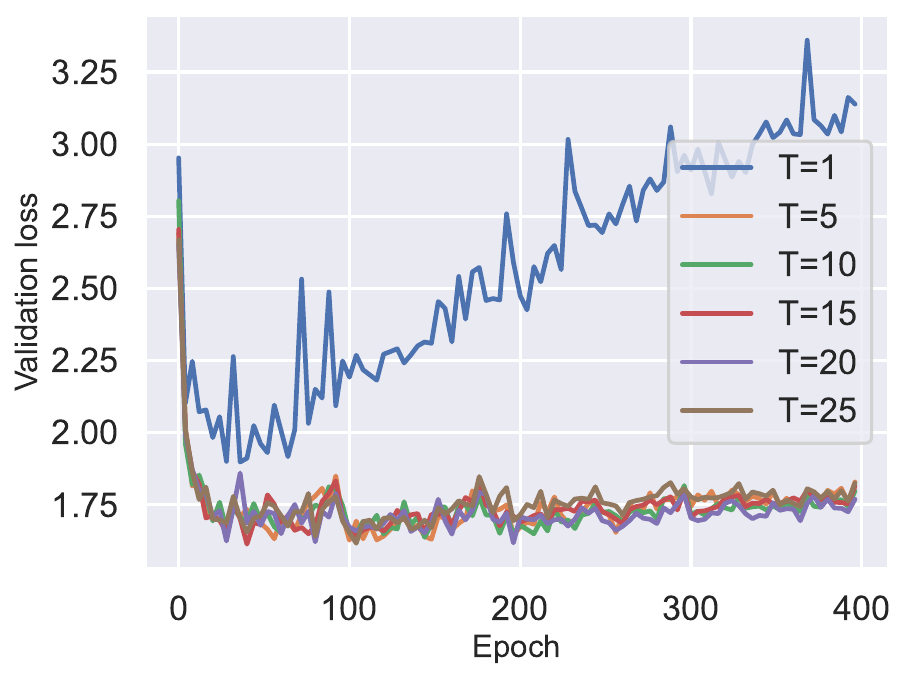}
         \caption{ResNet-50}
     \end{subfigure}
     \hfill
    \begin{subfigure}[t]{0.32\textwidth}
         \centering
         \includegraphics[trim = 0mm 0mm 0mm 0mm, clip, width=\textwidth]{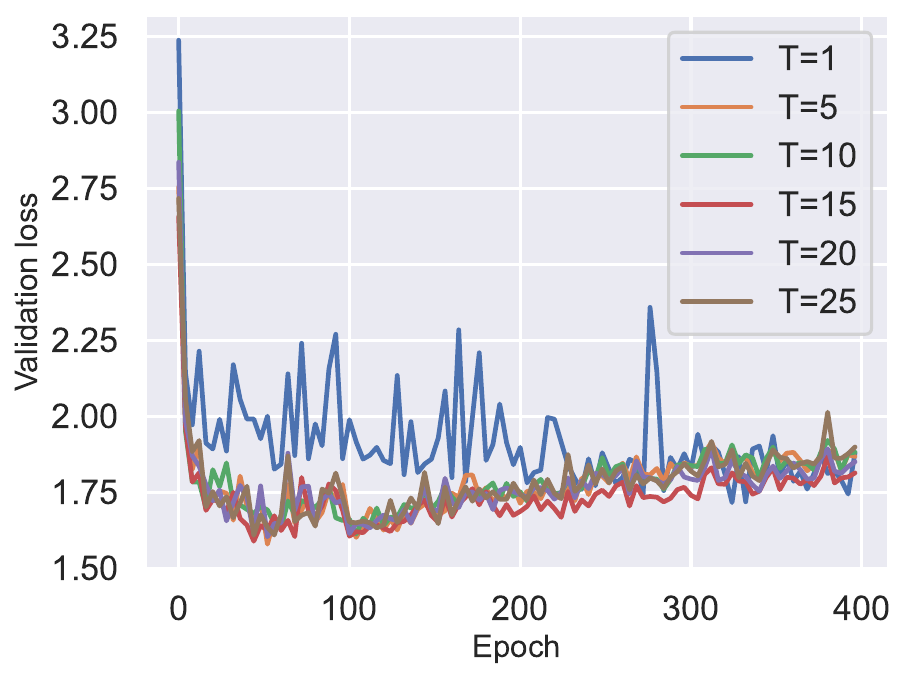}
         \caption{ResNet-101}
     \end{subfigure}
     \hfill
    \caption{Development of validation loss for $T$ values with initial step size 5 for NYUv2 using the SegNet, ResNet-50, and ResNet-101.
    }
    \label{fig:appx_searchT_rough_loss}
\end{figure}

In contrast, terminating non-optimal experiments based on the validation loss is not possible, as the different runs are all very similar in terms of their loss development, as can be seen in Figure \ref{fig:appx_searchT_rough_loss}.

\clearpage

\subsection{Performance across seeds}
\label{sec:appx_performance_seeds}
\begin{figure}
    \centering
    \includegraphics[scale=0.5]{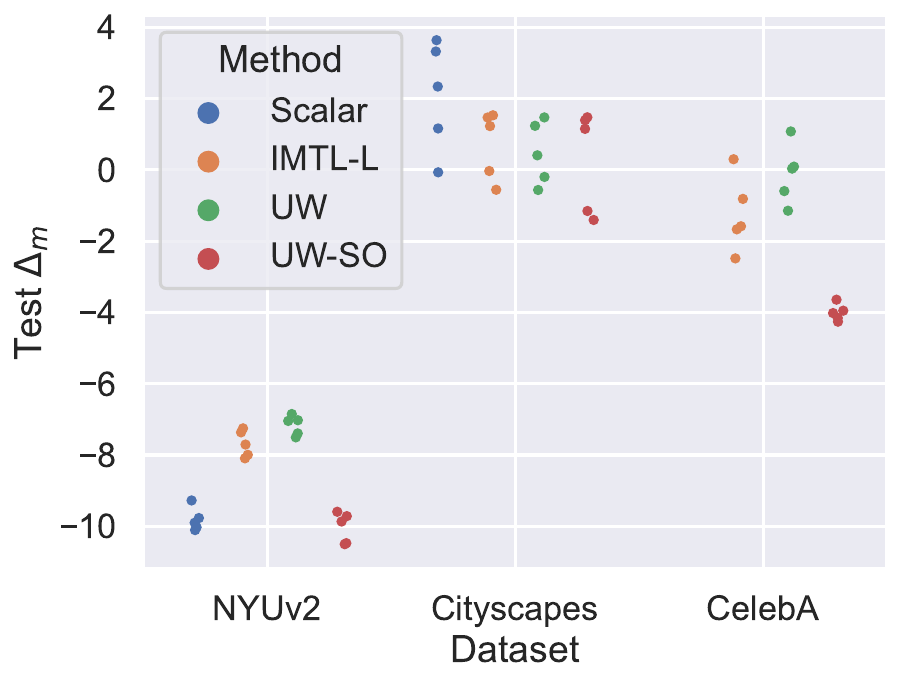}
    \caption{Test $\Delta_m$ metrics for 4 methods for NYUv2/Cityscapes (ResNet-50) and CelebA (ResNet-18) for 5 seeds. Even for the larger ResNet-50 where differences between methods become less extreme, UW-SO is almost always outperforming UW, except for 2 seeds on Cityscapes. Also, note the minimal variance on CelebA.}
    \label{fig:abl_seeds}
\end{figure}

\begin{figure}[ht]
    \centering
    \includegraphics[trim = 0mm 0mm 0mm 0mm, clip, width=\columnwidth]{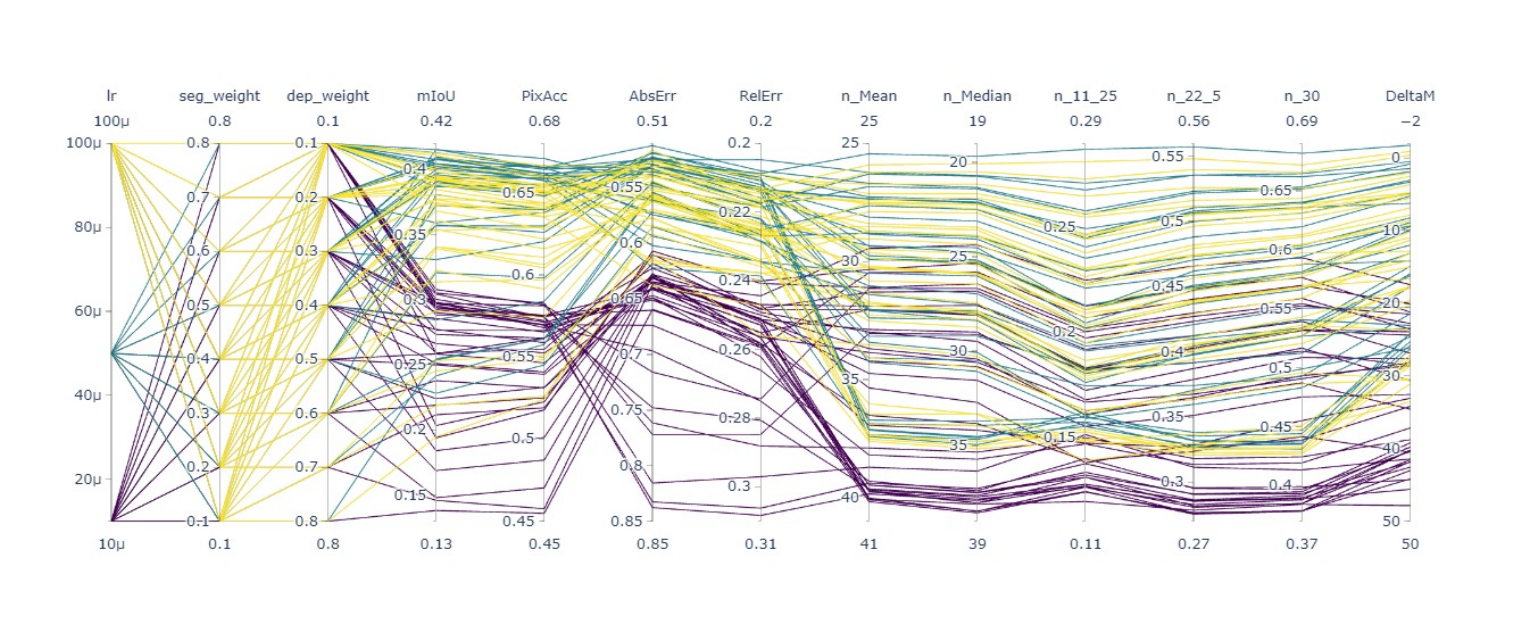}
    \caption{Results for performing the grid search of learning rate and task weights with a fixed weight decay of $10^{-5}$ using the Scalarization approach with the SegNet for the NYUv2 dataset. The learning rates are as follows: $10^{-4}$ (yellow), $5*10^{-5}$ (green), and $10^{-5}$ (purple). Each line represents one experiment for each individual combination of learning rate and task weights. Only 3 out of 108 combinations yield a $\Delta_m$ score $\leq 0$, meaning that only 3 combinations achieve a better average task performance than STL. Note that the metrics are ordered such that higher is always better (e.g., DeltaM is ordered from small to large while mIoU is ordered from large to small, thus, for both higher means better). The best run is given for $\omega_{sem} = 0.1$, $\omega_{dep} = 0.1$, $\omega_{norm} = 0.1$ with a learning rate of $5*10^{-5}$. Note that this is different from the final result in Table \ref{tab:nyuv2_results_main} as the weight decay is not yet tuned here.}
    \label{fig:parallel_coord_scalar}
\end{figure}

To show that UW-SO can be seen as a serious drop-in replacement of UW, we not only compare results across different backbone architectures but also across different network initializations (determined by the random seed). Figure \ref{fig:abl_seeds} depicts that UW-SO outperforms UW for almost any given random seed across all three datasets. It's important to emphasize that, in the case of NYUv2 and Cityscapes, we present results for the more proficient ResNet-50 model rather than the SegNet. This is done to underscore that the notable advantage of UW-SO persists even with larger architectures, where distinctions between MTO algorithms tend to be less pronounced.

\clearpage

\end{document}